\documentclass{article}

\usepackage{arxiv}

\usepackage{amsmath,amsfonts,bm}

\def\eqref#1{equation~\ref{#1}}
\def\floor#1{\lfloor #1 \rfloor}
\def\1{\bm{1}}

\DeclareMathAlphabet{\mathsfit}{\encodingdefault}{\sfdefault}{m}{sl}
\SetMathAlphabet{\mathsfit}{bold}{\encodingdefault}{\sfdefault}{bx}{n}

\newcommand{\R}{\mathbb{R}}

\DeclareMathOperator*{\argmax}{arg\,max}

\usepackage{algorithm}
\usepackage{algpseudocode}
\usepackage{listings}
\usepackage{comment}
\usepackage{bm}
\usepackage{caption}

\lstdefinestyle{prompt}{
  basicstyle=\ttfamily\scriptsize,
  breaklines=true,
  breakindent=0pt,
  columns=fullflexible,
  keepspaces=true,
  frame=single,
  framesep=4pt,
  xleftmargin=4pt,
  xrightmargin=4pt,
  aboveskip=6pt,
  belowskip=6pt,
}

\usepackage[utf8]{inputenc} % allow utf-8 input
\usepackage[T1]{fontenc}    % use 8-bit T1 fonts
\usepackage{hyperref}       % hyperlinks
\usepackage{url}            % simple URL typesetting
\usepackage{booktabs}       % professional-quality tables
\usepackage{amsfonts}       % blackboard math symbols
\usepackage{nicefrac}       % compact symbols for 1/2, etc.
\usepackage{microtype}      % microtypography
\usepackage{cleveref}       % smart cross-referencing
\usepackage{lipsum}         % Can be removed after putting your text content
\usepackage{graphicx}
\usepackage{natbib}
\usepackage{doi}

\newcommand{\nin}{\noindent}

\newcommand{\bea}{\begin{eqnarray}}
\newcommand{\eea}{\end{eqnarray}}
\newcommand{\beas}{\begin{eqnarray*}}
\newcommand{\eeas}{\end{eqnarray*}}
 
\newcommand{\ben}{\begin{enumerate}}
\newcommand{\een}{\end{enumerate}}

\newcommand{\midb}{\pmb{\mid}}

\newcommand{\ZZ}{\mathbb{Z}}
\newcommand{\RR}{\mathbb{R}}

\newcommand{\cD}{\mathcal{D}}

\newcommand{\cU}{\mathcal{U}}

\newcommand{\cN}{\mathcal{N}}

\newcommand{\cT}{\mathcal{T}}

\newcommand{\cG}{\mathcal{G}}
\newcommand{\p}{\vspace{-0.2cm}\paragraph}

\newif\ifshowcomments
\showcommentsfalse % <-- set to \showcommentsfalse to hide all comments after compilation
\ifshowcomments
  \newcommand{\vvb}[1]{{\small\sf \color{orange} VB: #1}} % Vadim Bulitko
  \newcommand{\sw}[1]{{\small\sf \color{olive} SW: #1}} % Shuwei Wang
  \newcommand{\rala}[1]{{\small\sf \color{cyan} RL: #1}} % Ramon Lawrence
  \newcommand{\mb}[1]{{\small\sf \color{yellow} MB: #1}} % Matthew Brown
  \newcommand{\wy}[1]{{\small\sf \color{brown} WY: #1}} % William Yeoh
  \newcommand{\sn}[1]{{\small\sf \color{purple} SN: #1}} % Shinichi Nakagawa
  \newcommand{\my}[1]{{\small\sf \color{teal} MY: #1}} % Michael Youngblood
  \newcommand{\yw}[1]{{\small\sf \color{magenta} YW: #1}} % Yu Wang
\else
  \newcommand{\vvb}[1]{} \newcommand{\sw}[1]{} \newcommand{\rala}[1]{}
  \newcommand{\mb}[1]{} \newcommand{\wy}[1]{} \newcommand{\sn}[1]{}
  \newcommand{\my}[1]{} \newcommand{\yw}[1]{}
\fi

\title{Synthesizing Physics Formulae with Transformers}

\usepackage{authblk}

\author[1]{Shuwei Wang}
\author[1]{Vadim Bulitko}
\author[4]{Michael Youngblood}
\author[5]{Ramon Lawrence}
\author[3]{William Yeoh}
\author[2]{Shinichi Nakagawa}
\author[1]{Matthew R.~G.~Brown}
\author[1]{Yu Wang}
\affil[1]{Department of Computing Science, University of Alberta, 116 Street and 85 Avenue, Edmonton, AB, T6G 2R3, Canada}
\affil[2]{Department of Biology, University of Alberta, 116 Street and 85 Avenue, Edmonton, AB, T6G 2R3, Canada}
\affil[3]{McKelvey School of Engineering, Washington University in St.\ Louis, 1 Brookings Drive, St.\ Louis, MO 63130-4899, USA}
\affil[4]{Filuta AI, 1606 Headway Circle, Austin, TX 78754, USA}
\affil[5]{Department of Computer Science, Data Science, University of British Columbia, 6200 University Blvd, Vancouver, BC Canada V6T 1Z4, Canada}
\affil[ ]{\texttt{\{shuwei4,bulitko\}@ualberta.ca}, \texttt{gmyoungblood@gmail.com}, \texttt{ramon.lawrence@ubc.ca}, \texttt{william.yeoh@gmail.com}, \texttt{snakagaw@ualberta.ca}, \texttt{mrbrown23@gmail.com}, \texttt{yu35@ualberta.ca}}

\renewcommand{\shorttitle}{Synthesizing Physics Formulae with Transformers}

\hypersetup{
pdftitle={A template for the arxiv style},
pdfsubject={q-bio.NC, q-bio.QM},
pdfauthor={Shuwei Wang, Vadim Bulitko, Michael Youngblood, Ramon Lawrence, William Yeoh, Shinichi Nakagawa, Matthew R. G. Brown, Yu Wang},
pdfkeywords={Symbolic regression, Transformers, Equation discovery, Extrapolation, Training data generation, Formula simplification},
}

\begin{document}
\maketitle

\begin{abstract}
Finding a compact formula that fits a set of input-output pairs and predicts outputs on unseen inputs is a fundamental problem in science. Symbolic regression automates the search for such formulae: search-based methods explore the space of possible formulae directly, while transformers pre-trained on synthetic data produce formulae of comparable quality substantially faster. Existing transformers, however, are prone to overfitting --- they find formulae that fit the training data well but do not extrapolate to input ranges unseen during training. We address this by shaping the set of formulae used to train a transformer, and show that the resulting formulae extrapolate substantially better. Fine-tuning the transformer on data with noise-corrupted target values further makes the synthesized formulae robust to noise in the observations. On SRBench and LLM-SRBench our transformer synthesizes a formula in about ten seconds and extrapolates better than all evaluated methods at a comparable budget. Search-based methods surpass our accuracy only when given one to three orders of magnitude more time.
\end{abstract}

% keywords can be removed

%%%%%%%%%%%%%%%%%%%%%%%%%%%%%%%%%%%%%%%%%%%%%%%%%%%%%%%%%%%%

\section{Introduction}
\label{sec:introduction}

Symbolic regression seeks to produce a formula that fits a set of observations~\citep{schmidt2009distilling, lample2020deep}. Genetic algorithms (GAs) search the space of formulae seeking one that fits the data the most. While GAs can find formulae with a high quality of fit~\citep{udrescu2020aifeynman} the process can be slow~\citep{lacava2021contemporary}. A recent line of work pre-trains transformers to produce symbolic formulae based on input observations. As real-life observations are often limited in volume, the transformers are trained on synthetic observations produced by symbolic formulae randomly drawn from a given grammar. This allows one to readily produce large and diverse sets of training data, each labeled with the generated formula. Once trained the transformers can produce symbolic formulae fitting test-time observations with the quality of fit comparable with state-of-the-art GA methods while doing so orders of magnitude faster~\citep{biggio2021nesymres, kamienny2022e2e, shojaee2023tpsr}. A fast formula synthesis can be useful for practical applications with time constraints such as programmatic reinforcement learning~\citep{kubalik2021symbolic, derner2020constructing}. The downside of using existing pre-trained transformers is that they often produce complex symbolic formulae that fit the training data well but fail to extrapolate, limiting the use of the synthesized symbolic formulae in practice~\citep{voigt2025analyzing}.

We address this problem by simplifying each randomly drawn formula into an equivalent but more compact and consistently ordered form and using that form as the training target. We show that a transformer trained on such targets produces more compact symbolic formulae that extrapolate substantially better to unseen input ranges than existing formula-trained transformers. This allows the formulae produced by our transformers to serve as predictive models rather than mere data compressors~\citep{schmidhuber2008driven, segui2026geometric, schulte2009simultaneous}.

Our empirical work compares our method with genetic-algorithm, formula-trained transformer and language-model methods on SRBench~\citep{lacava2021contemporary} and LLM-SRBench~\citep{shojaee2025llmsrbench}. Our transformer synthesizes a formula in about ten seconds, and at that cost it extrapolates better than every other method we evaluate that takes under a minute per formula. Search-based methods extrapolate better when given one to three orders of magnitude more time: PySR with a $30$-minute budget on SRBench, and AI Feynman on noise-free SRBench, and PySR with a $30$-minute budget and LLM-SR with Gemini-3.5-flash on LSR-Transform. On the LSR-Synth tasks, whose inputs lie at physical scales far from those of our training data, our transformer trails search-based methods. Our code\footnote{\url{https://github.com/ShwayW/shorter-formulae}} and model weights\footnote{\url{https://huggingface.co/ShwayW/shorter-formulae}} are publicly available.

%%%%%%%%%%%%%%%%%%%%%%%%%%%%%%%%%%%%%%%%%%%%%%%%%%%%%%%%%%%%

\section{Problem Formulation}
\label{sec:problemFormulation}

Consider a function $ f : \mathbb{R}^m \to \mathbb{R} $. The problem is to synthesize a formula $ \tilde{f} \in F $ which approximates $ f $ over a finite set of inputs $ D \subset  \mathbb{R}^m $. The quality of approximation is measured via the coefficient of determination:
\beas
	R^2(f,\tilde{f},D) &= 1 - \frac{\sum_{\mathbf{x}_i \in D}\left(f(\mathbf{x}_i) - \tilde{f}(\mathbf{x}_i)\right)^2}{\sum_{\mathbf{x}_i \in D}\left(f(\mathbf{x}_i) - \bar{f}\right)^2} 
\eeas
\nin where $ \bar{f} $ is the mean value of $ f $ over the inputs $ \mathbf{x}_i \in D $. The input to the synthesizer is a training set $ D_\text{train} = \{(\mathbf{x}_i, f(\mathbf{x}_i))\}^n_{i = 1} $ with $ f $ the target function. The performance of the synthesized approximation is measured on $ D_\text{test} $ and the task is to find $ f^* $ to maximize the coefficient of determination:
\beas
	f^* &= \argmax_{\tilde{f} \in F} R^2(f,\tilde{f},D_\text{test}).
\eeas
$ D_\text{test} $ is assumed to be different from $ D_\text{train} $ but the ground-truth $ f $ should fit it exactly: $ R^2(f, f, D_\text{test}) = 1 $. We prefer synthesis to have low computational cost and to produce compact formulae.

To measure extrapolation we shift the input range of $ D_\text{train} $ by $ k $ times its width:
\beas
	D^k_\text{test} &= \left\{(\mathbf{x}_i, f(\mathbf{x}_i)) \mid \mathbf{x}_i \sim \cU(\bm{\ell} + k \cdot \boldsymbol{\delta}, \mathbf{r} + k \cdot \boldsymbol{\delta})\right\}^n_{i = 1}
\eeas
where $ \bm{\ell} = \min_{\mathbf{x} \in D_\text{train}}(\mathbf{x}) $, $ \mathbf{r} = \max_{\mathbf{x} \in D_\text{train}}(\mathbf{x}) $ element-wise and $ \boldsymbol{\delta} = \mathbf{r} - \bm{\ell} $. Here $ \cU(\bm{\ell} + k \cdot \boldsymbol{\delta}, \mathbf{r} + k \cdot \boldsymbol{\delta}) $ is a multivariate uniform distribution over the hyper-rectangle $ [\bm{\ell} + k \cdot \boldsymbol{\delta}, \mathbf{r} + k \cdot \boldsymbol{\delta}] $. We define the degree-$ k $ $ R^2 $ denoted by $ R^2_k(f, \tilde{f}) $ as the coefficient of determination evaluated on $ D^k_\text{test} $.

We report, for each $ k $, the fraction of tasks on which the synthesized formula attains a near-exact fit on $ D^k_\text{test} $. Let $ \cT $ be a benchmark of synthesis tasks, with $ f_t $ the target formula of task $ t \in \cT $ and $ {\tilde f}_t $ synthesized for it, we define the degree-$ k $ solution rate as
\beas
      \rho_k(\cT) &= \frac{1}{|\cT|} \left| \left\{ t \in \cT \mid R^2_k(f_t, {\tilde f}_t) \ge \tau \right\} \right|
\eeas
where $ \tau $ is for a near-exact match. We also report the mean synthesis time per task and the mean {\em size} of a synthesized formula, defined as the number of nodes in it.

%%%%%%%%%%%%%%%%%%%%%%%%%%%%%%%%%%%%%%%%%%%%%%%%%%%%%%%%%%%%

\section{Related Work}
\label{sec:relatedWork}

Symbolic regression methods applicable to the problem defined in the previous section fall into three categories. {\em Search-based symbolic regression}~\citep{schmidt2009distilling, cranmer2020discovering, petersen2021deep, bulitko2022portability, wang2023synthesizing, sun2023symbolic, cranmer2023interpretable} uses genetic algorithms or reinforcement learning to search the space of formulae. Such methods do not require pre-training but are typically computationally expensive.

{\em Symbolic regression based on large language models}~\citep{grayeli2024symbolic, ma2024llm, hua2025finetuning, shojaee2025llmsr} involves training on a large corpus of text in natural languages (and other data) which then allows it to use knowledge of physics to synthesize formulae. During synthesis the large language models can be run for inference repeatedly within a search-based algorithm, trading synthesis time for fit accuracy. However, the performance of these methods depends on the quality of the underlying language models.

Instead of training a general large language model, one can also train a smaller {\em transformer on the input-output pairs and the formulae only}, without any natural language data~\citep{biggio2021nesymres, valipour2022symbolicgpt, kamienny2022e2e, shojaee2023tpsr, d2024odeformer, vastl2024symformer, ying2025neural}. Doing so speeds up and simplifies the training but during synthesis such transformers tend to produce formulae that overfit the input data, exhibiting poor extrapolation~\citep{voigt2025analyzing}. To improve accuracy of synthesized formulae on inputs outside of the training domain, researcher have attempted to shape the distribution of the data on which the transformer is trained~\citep{biggio2021nesymres, bendinelli2023controllable,saegert2026simplipy}. Shaping usually rewrites each generated formula with SymPy~\citep{meurer2017sympy}, a throughput bottleneck when formulae are generated online that~\citet{saegert2026simplipy} replace with offline-mined rewrite rules applied by pattern matching. We instead keep one formula per equivalence class, with no redundant constants for overfitting to the training range.

%%%%%%%%%%%%%%%%%%%%%%%%%%%%%%%%%%%%%%%%%%%%%%%%%%%%%%%%%%%%

\section{Proposed Approach}
\label{sec:proposedApproach}

We train an encoder-decoder transformer such that given $ N $ data points $ D = \{(\mathbf{x}_i, y_i)\}^N_{i = 1} $ where for each $ i ,\ f(\mathbf{x}_i) = y_i $ with $ f : \R^m \to \R $ the target formula, the transformer receives $ D $ and synthesizes $ {\tilde f} $ token by token. Afterwards the constants are fitted to $ D $.

\subsection{Training Data Generation}
\label{subsec:dataGeneration}

We define a space of formulae $ \cG $ by the following context-free grammar~\citep{udrescu2020ai}:
{\small\beas
S &\to& T \midb U \midb B \nonumber \\
T &\to& v_1 \midb v_2 \midb v_3 \midb v_4 \midb v_5 \midb v_6 \midb v_7 \midb v_8 \midb v_9 \midb v_{10} \midb C \nonumber  \\
U &\to& S+\!+ \midb S-\!- \midb -S \midb \sqrt{S} \midb e^S \midb \ln(S) \midb \sin(S) \midb \nonumber \cos(S) \midb \tan(S) \midb |S| \midb \arcsin(S) \midb \arccos(S) \nonumber \\
& & \arctan(S) \midb \frac{1}{S} \midb  \tanh(S) \midb S^2 \midb S^3 \nonumber \\
B &\to& S+S \midb S-S \midb S \times S \midb \frac{S}{S} \midb S^S
\eeas}
where $ C $ is a set of constants.

During training, each data instance is of the form $ (f, D) $ where $ f $ is a target formula from the space $ \cG $ in standard Polish notation~\citep[a prefix notation where operators come before their operands]{biggio2021nesymres} and $ D $ is a set of data points from the space $ \RR^{N \times (v + 1)} $ where $ v $ is the number of variables in $ f $ with the last column of $ D $ the outputs of $ f $ given the $ v $ inputs. A formula can be seen as a tree with the first operator the root node and the variables and constants as leaves.

\p{Formulae generation.} To create a formula for training we sample the number of variables $ v $ from a uniform distribution $ \cU(1, V_\text{max}) $. We then sample the total number of nodes $ n $ of the formula from $ \cD(l, r, \gamma) $, a distribution over $ [l, r] $ with decay $ \gamma \in (0, 1] $ that biases the training data towards shorter formulae. We then sample the number of unary operators $ u $ from $ \cD(0, \min(8, \max(0, \floor{(n - 1) / 2})), \gamma) $ with the number of binary operators following $ n $ and $ u $.

We then generate a formula skeleton with $ v $ variables, $ n $ nodes and $ u $ unary operators following the method of~\citet{lample2020deep} as adapted by~\citet{kamienny2022e2e}. The distribution over the set of binary operators and unary operators are set by hand (Appendix~\ref{app:operatorWeights}). Once a formula skeleton is generated we sample from a log-uniform distribution~\citep{matsubara2024rethinking, d2024odeformer} for each constant $ c = s \cdot 10^q $ with $ s \sim \cU\{-1, 1\} $ and $ q \sim \cU(-3, 3) $. 

Existing formula-trained transformers place fittable constants at the additive positions of a skeleton~\citep{kamienny2022e2e}: each term of a sum is multiplied by a constant, the argument $ S $ of a unary operator is replaced by $ a + b \cdot S $ (unless that argument is itself a sum) and the formula becomes $ c_0 + f $. Products and powers receive no constants of their own, so $ v_1 \times v_2 $ is drawn as $ c_0 + c_1 v_1 v_2 $, while $ \sin(v_1) $ is drawn as $ c_0 + \sin(c_1 + c_2 v_1) $. Our generator does not apply it.

\p{Formula simplification.} Many different formulae are functionally equivalent. A full standardization is undecidable in general~\citep{richardson1968}, so we do not attempt it. In the spirit of~\citet{cohen2003computer} we flatten each chain of connected additive operators: $ S + S $, $ S - S $, and the following unary operators $ S+\!+ $, $ S-\!- $, $ -S $, into a sum $ c_0 + \sum_i c_i t_i $ with like terms collected. Multiplicative chains $ S \times S $, $ S / S $, $ \frac{1}{S} $ and the powers $ S^n $ with integer $ n $ are flattened the same way into the product $ c \prod_i b_i^{e_i} $ with like factors collected. We merge exponents only when they are integers, which is value-safe as $ S^m S^n = S^{m+n} $ for integers $ m, n $ whenever $ S \neq 0 $. We call a subformula an {\em atom} when the flattening treats it as a single indivisible term or factor and rewrites only its argument. We keep $ \sqrt{S} $, $ |S| $, $ S^k $ with non-integer $ k $, and the transcendental unary operators $ e^S $, $ \ln(S) $, $ \sin(S) $, $ \cos(S) $, $ \tan(S) $, $ \arcsin(S) $, $ \arccos(S) $, $ \arctan(S) $ and $ \tanh(S) $ as atoms, and flatten their arguments recursively. We also apply the cancellations $ \ln e^S = S $, $ -(-S) = S $, $ ||S|| = |S| $, $ e^{\ln S} = S $ and $ 1/(1/S) = S $. We then sort the resulting sums and products into a fixed order and re-index the variables in ascending order, so that $ v_3 + v_2 \cdot v_4 - v_1 $ becomes $ v_1 + v_2 \cdot v_3 - v_4 $. Appendix~\ref{app:simplification} gives the full ordering rules, the effect of the rewrites on the domain, and the pseudocode for the flattening.

\p{Data points generation.} Once a formula is simplified by the method above we sample its inputs as~\citet{kamienny2022e2e} do, from a mixture of $1$ to $10$ Gaussian or uniform clusters with centers drawn from $\cN(\mathbf{0}, I)$, a scale per variable drawn from $\cU(0,1)$ and a random rotation per cluster, and calculate $ y_i = f(\mathbf{x}_i) $. We discard inputs at which $ f $ is undefined or overflows ($ |f(\mathbf{x}_i)| \ge 10^9 $) and draw a fresh mixture for the missing points until $ N $ valid data points remain. We whiten the inputs, $ \mathbf{\tilde x}_i = (\mathbf{x}_i - \boldsymbol{\mu}) / \boldsymbol{\sigma} $ with $ \boldsymbol{\mu} $ and $ \boldsymbol{\sigma} $ the per-variable mean and standard deviation over the $ N $ data points, so every input the transformer receives has zero mean and unit variance.

We whiten the inputs during both training and synthesis, but not the outputs. Unlike~\citet{kamienny2022e2e}, we compute each $ y_i $ from the raw inputs $ \mathbf{x}_i $, so the transformer learns $ f $ on the original inputs.

\subsection{Transformer Architecture and Training}
\label{subsec:transformerArchitectureTraining}

We introduce our embedding method for the data points as well as our transformer's architecture. We then detail the training process.

\p{Tokenization and embedding.} We round each floating point number to four significant digits~\citep{kamienny2022e2e} and encode it as four tokens~\citep{charton2022linear}: the sign ($-$ or $+$), the higher mantissa ($00$ to $99$), the lower mantissa ($00$ to $99$), and the exponent ($-100$ to $100$), a vocabulary of size $2 + 100 + 201 = 303$. Variables absent from the formula are padded with a dedicated token \texttt{<INPUT\_PAD>}. We embed and flatten the tokens, then a two-layer ReLU multi-layer perceptron maps them to a $d_\text{emb}$-dimensional input of the transformer encoder. 

\p{Transformer architecture.} We use an encoder-decoder transformer architecture~\citep{vaswani2017attention, phuong2022formal}. Since the order of data points does not matter to the synthesized formulae we remove the positional embedding from the encoder~\citep{biggio2021nesymres, kamienny2022e2e}. Starting from the $ 89 $ million-parameter model proposed by~\citet{kamienny2022e2e}, we enlarge our transformer to have $ 12 $ encoder layers and $ 20 $ decoder layers. We set an embedding dimension of $ 512 $ and $ 20 $ attention heads with $ 32 $-dimensional queries, keys and values, for a total of about $ 145 $ million parameters. To stabilize training we use the Pre-LayerNorm~\citep{xiong2020layer} architecture and the GELU activation function~\citep{hendrycks2016gaussian} in the transformer.

\p{Training.} We use the Adam~\citep{kingma2015adam} optimizer to minimize the cross-entropy loss, increasing the learning rate linearly from $ 1.0 \times 10^{-7} $ to $ 4.0 \times 10^{-4} $ over the first $ 10{,}000 $ training steps and then lowering the learning rate by inverse-square-root down to $ 1.0 \times 10^{-4} $. We use $ 18.2 $k generated data instances (equivalent to $ 200 $ batches) as validation data and train the transformer with a patience of $ 10 $ epochs for early stopping. To avoid wasteful padding we batch together data instances with formulae of similar lengths. There are approximately $ 1.1 $M tokens passing through the encoder for each full batch. We train our transformer on a single NVIDIA H100 GPU with 80 GB memory. A single epoch takes about 30 minutes and the full training about two weeks. The values of all the control parameters of our data generation, architecture, training and synthesis are collected in Appendix~\ref{app:controlparameters}.

\p{Fine tuning for noisy targets.} We further train the transformer on noisy targets: for each formula we draw a noise level $ \epsilon \sim \cU(0, 0.1) $ in the convention of Section~\ref{subsec:control_parameters}, and set $ y \leftarrow y + \epsilon \sqrt{\overline{y^2}} \xi $ with $ \xi \sim \cN(0, 1) $, holding half the formulae noise-free so the noise-free objective is not forgotten. We resume from the trained transformer with a peak learning rate of $ 5 \times 10^{-5} $.

\subsection{Inference Methods}
\label{subsec:inferenceMethods}

We compare beam search~\citep{wiseman2016sequence} with two hybrid inference methods for our transformer: transformer-based planning for symbolic regression (TPSR)~\citep{shojaee2023tpsr} and the divide-and-conquer strategy (D\&C)~\citep{ying2025neural}.

\p{Whitened and raw variables.} Our transformer reads whitened inputs, so the formula it synthesizes is a function of $ \mathbf{\tilde x} $. Following~\citet{kamienny2022e2e} one can substitute $ v_i $ with $ (v_i - \mu_i)/\sigma_i $ to express it in the original variables, which makes the synthesis insensitive to the scale of the inputs. We do not do so by default. The substitution writes the location and scale of $ D_\text{train} $ into the formula as constants, and those constants are wrong once the inputs move away from the range they were computed on. It also replaces every occurrence of a variable with a subexpression, inflating the size of the synthesized formula. We instead keep the formula as the transformer emits it and fit its constants against the raw data points with BFGS, and we report the substituted variant separately as {\em ours (unscaled)}.

\p{Bagging and refinement.} Since our transformer is trained on at most $ N $ data points, whenever there are more than $ N $ data points available for a task we draw $ B $ bags of $ N $ data points each, sampling uniformly with replacement so the bags are independent but not a partition of the available points. We use the synthesized constants as initial values and refine them {\em a posteriori} with the Broyden–Fletcher–Goldfarb–Shanno (BFGS) algorithm~\citep{fletcher2013practical}. We keep the top $ C $ candidate formulae with beam search giving us in total $ BC $ candidate formulae.

\p{Transformer-based planning.} Following~\citet{shojaee2023tpsr}, the next-token distribution of our transformer serves as the prior of a Monte-Carlo tree search. Beam search completes each expansion into a formula $ \hat f $, and the reward $ 1/(1+\mathrm{NMSE}) + \lambda e^{-|\hat f|/200} $ on $ D_\text{train} $ is propagated back up the tree.

\p{Divide and conquer.} We run beam search on the original task first. If its formula reaches $ R^2 \ge \theta $ on a held-out split of $ D_\text{train} $ we keep it, otherwise we decompose. We fit a feed-forward network to $ D_\text{train} $, one per task and from scratch, as a differentiable proxy for the unknown function (Appendix~\ref{app:oracleNetwork}). Following~\citet{ying2025neural} we read the structure of $ f $ off its second derivatives, testing additive and multiplicative separability only. Two variables are additively separable when $ \partial^2 f / \partial x_i \partial x_j $ vanishes and multiplicatively separable when $ \partial^2 \ln|f| / \partial x_i \partial x_j $ vanishes. Since $ \partial^2 \ln|f| / \partial x_i \partial x_j = H_{ij}/f - (\partial_i f)(\partial_j f)/f^2 $, one gradient and Hessian of the network give both tests. We take the median of each over the sample points to form a coupling matrix, threshold it at several cutoffs, and group the variables by the connected components of the pairs that stay coupled. For each grouping we query the network for data points on every group, holding the variables outside it at their median. Our transformer synthesizes a subformula per group, and we combine them into one formula whose constants we refine with BFGS against $ D_\text{train} $. We keep the beam-search formula unless a combined one scores higher on the held-out split of $ D_\text{train} $.

%%%%%%%%%%%%%%%%%%%%%%%%%%%%%%%%%%%%%%%%%%%%%%%%%%%%%%%%%%%%

\section{Empirical Evaluation}
\label{sec:empiricalEvaluation}

We introduce the two benchmarks we use for evaluation in Section~\ref{subsec:benchmarks} and go through the control parameters in Section~\ref{subsec:control_parameters} and then show our evaluation results in Section~\ref{subsec:results}.

\subsection{Benchmarks}
\label{subsec:benchmarks}

We evaluate the methods using two benchmarks: the SRBench~\citep{lacava2021contemporary}, from which we take the $ 99 $ Feynman formulae with ground-truth formulae available to compute the extrapolation set $ D^k_\text{test} $ at every $ k $, and the LSR-Transform category of the LLM-SRBench~\citep{shojaee2025llmsrbench} which provides 111 ground-truth formulae. Both benchmarks have no more than ten variables. We also evaluate our method on the LSR-Synth category of the LLM-SRBench~\citep{shojaee2025llmsrbench} which contains 129 formulae.

\subsection{Control Parameters}
\label{subsec:control_parameters}

For each task we construct $ D^k_\text{test} $ as in Section~\ref{sec:problemFormulation}, with $ n = 10{,}000 $ points, for $ k \in \{0, 1, 2, 4, 8, 16, 32, 64, 128\} $. Shifting every variable by $ k \boldsymbol{\delta} $ can create a domain of the target formula where it is undefined so we advance each variable as far as it can go and freeze the rest, keeping the formulae defined for every $ k $ (Appendix~\ref{app:ooddomain}). At $ k = 0 $, $ D^0_\text{test} $ is a sample over the range of $ D_\text{train} $. We set the near-exact match threshold to $ \tau = 0.99 $ following existing research~\citep{kamienny2022e2e, shojaee2023tpsr}. To test robustness of the methods under noise we corrupt the target values of $ D_\text{train} $ with additive Gaussian noise, $ y \leftarrow y + \xi $ with $ \xi \sim \cN(0, \sigma^2) $ and $ \sigma = \epsilon \sqrt{\overline{y^2}} $, where $ \overline{y^2} $ is the mean squared target value over $ D_\text{train} $, with the set of target noise levels $ \epsilon \in \{0.0, 0.001, 0.01, 0.1\} $. The target values of every $ D^k_\text{test} $ are left noise-free. We run each method with $ 10 $ random number-generator seeds and plot the means and the standard deviations over runs.

\p{Data budget.} For each task and run we draw a pool of data points uniformly at random from those the benchmark provides and split it $ 75/25 $ into $ D_\text{train} $ and a validation set, drawing the $ B $ bags from $ D_\text{train} $ as described in Section~\ref{subsec:inferenceMethods}. On SRBench the pool holds $ \lceil BN/0.75 \rceil = 53{,}334 $ points, so $ D_\text{train} $ holds $ BN = 40{,}000 $ points. On LLM-SRBench we use the training and test splits the benchmark provides and cap the pool at $ 20{,}000 $ points. We run beam search and TPSR fit and select on all of them, while the D\&C strategy takes the same $ 75/25 $ split. The validation set is used only to select among candidate formulae during synthesis.

\subsection{Results}
\label{subsec:results}

We compare our method with search-based methods, formula-trained transformers and LLMs, and read the comparison as a trade-off between the solution rate $ \rho_k $ and the synthesis time.

\begin{figure*}[t]
\centering
\includegraphics[width=\textwidth]{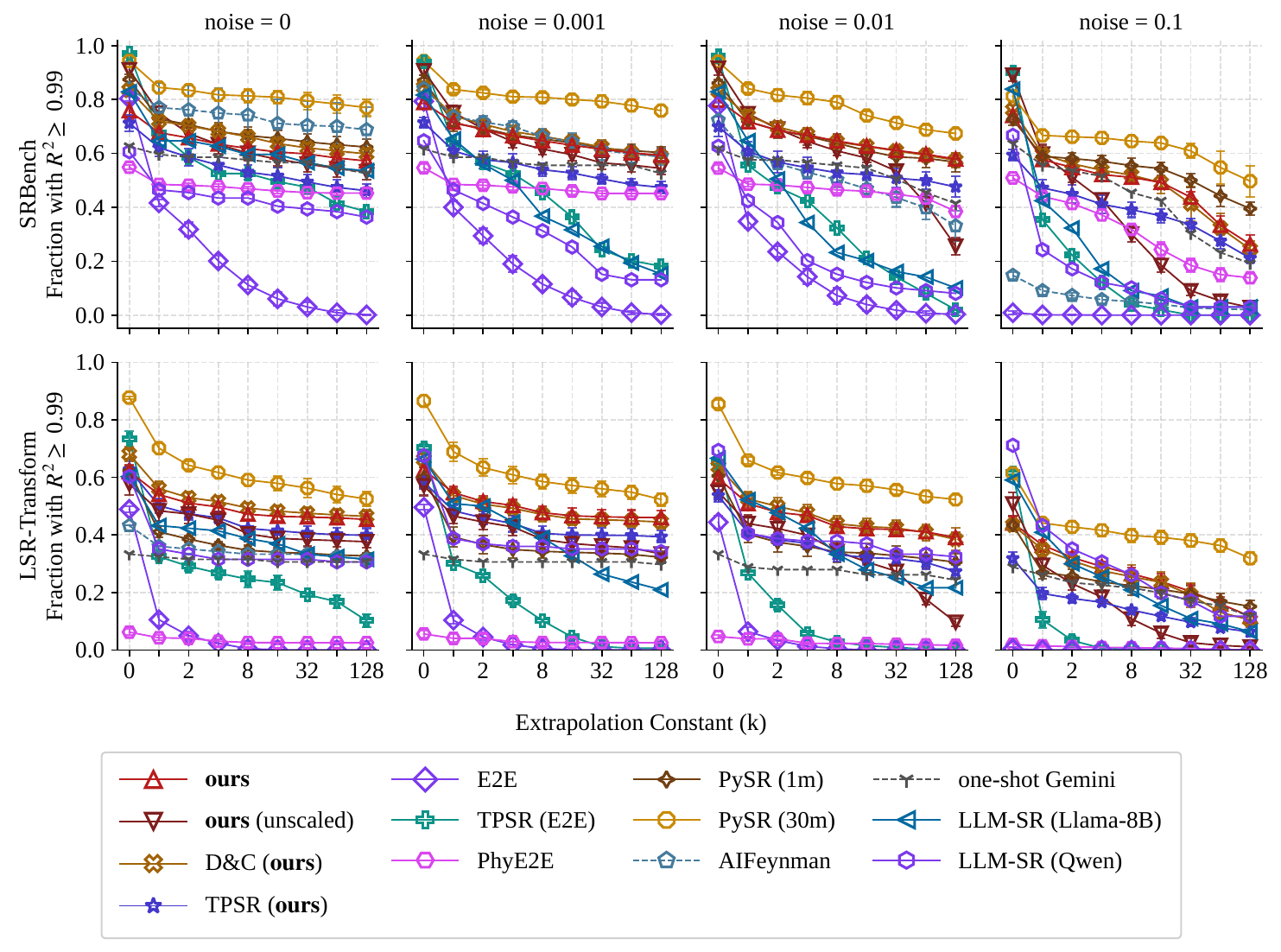}
\caption{Solution rate $ \rho_k $ versus extrapolation constant $ k $ on the $ 99 $ SRBench formulae and on the $ 111 $ LSR-Transform formulae.}
\label{fig:both_extrapolation}
\end{figure*}

\p{SRBench methods.} \citet{lacava2021contemporary} ran $ 14 $ symbolic regression methods on the SRBench tasks and released the formula each method synthesized. We evaluate these released formulae on our $ D^k_\text{test} $ and show the method with the highest $ \rho_k $, AI Feynman~\citep{udrescu2020ai}.

\p{PySR.} We also evaluate PySR~\citep{cranmer2023interpretable}, a recent search-based method involving a genetic algorithm. We keep the operator set, the constraints on operator nesting and the population settings the SRBench specifies, with a largest formula size of $ 30 $ and a largest depth of $ 20 $. We run it at four time budgets per formula, $ 10 $ seconds, $ 1 $ minute, $ 30 $ minutes and $ 1 $ hour, and report them as {\em PySR (10s)}, {\em PySR (1m)}, {\em PySR (30m)} and {\em PySR (1h)}.  Each fit runs on a single CPU core, and the measured time includes the start-up of PySR's Julia back end (about $ 40 $ seconds). The main figures show PySR (30m) only, since PySR (1h) differs from it by at most $ 0.02 $ in $ \rho_k $.

\p{Formula-trained transformers.} We evaluate six formula-trained transformer methods: E2E~\citep{kamienny2022e2e}, TPSR~\citep{shojaee2023tpsr}, PhyE2E~\citep{ying2025neural}, NeSymReS~\citep{biggio2021nesymres}, SymFormer~\citep{vastl2024symformer}, and TF4SR~\citep{lalande2023transformer}, and show the three best, E2E, TPSR and PhyE2E, here (all six in Appendix~\ref{app:fml_transformer_arms}). We use E2E with beam search. TPSR decodes with the pre-trained E2E model using Monte Carlo tree search guided by a reward that combines the quality of fit and the complexity of the formula. PhyE2E trains a similar architecture to E2E on LLM-generated data and applies D\&C but we evaluated it using beam search.

\p{Large language models.} We evaluate three open-source models, Llama-3.1-8B-Instruct (Llama-8B), Llama-3.3-70B-Instruct (Llama-70b)~\cite{grattafiori2024llama} and Qwen-2.5-Coder-32B (Qwen)~\cite{hui2024qwen2}, and one commercial model, Gemini-3.5-flash (Gemini), on the two benchmarks (Appendix~\ref{app:llmarms}). We evaluate the LLMs in two ways: a one-shot method that prompts the LLM once for a formula skeleton and then applies BFGS to fit the constants to the training data (see Appendix~\ref{app:oneshot-prompt} for the prompt) and LLM-SR~\citep{shojaee2025llmsr} which uses the LLM to generate an initial population of candidate formulae and iteratively refine them based on loss. We set a budget limit of $ 1{,}000 $ LLM calls per LLM-SR run following previous work~\citep{shojaee2025llmsrbench}. They are not restricted to the operators of $ \cG $ (Appendix~\ref{app:operatorSets}).

\begin{figure*}[t]
\centering
\includegraphics[width=\textwidth]{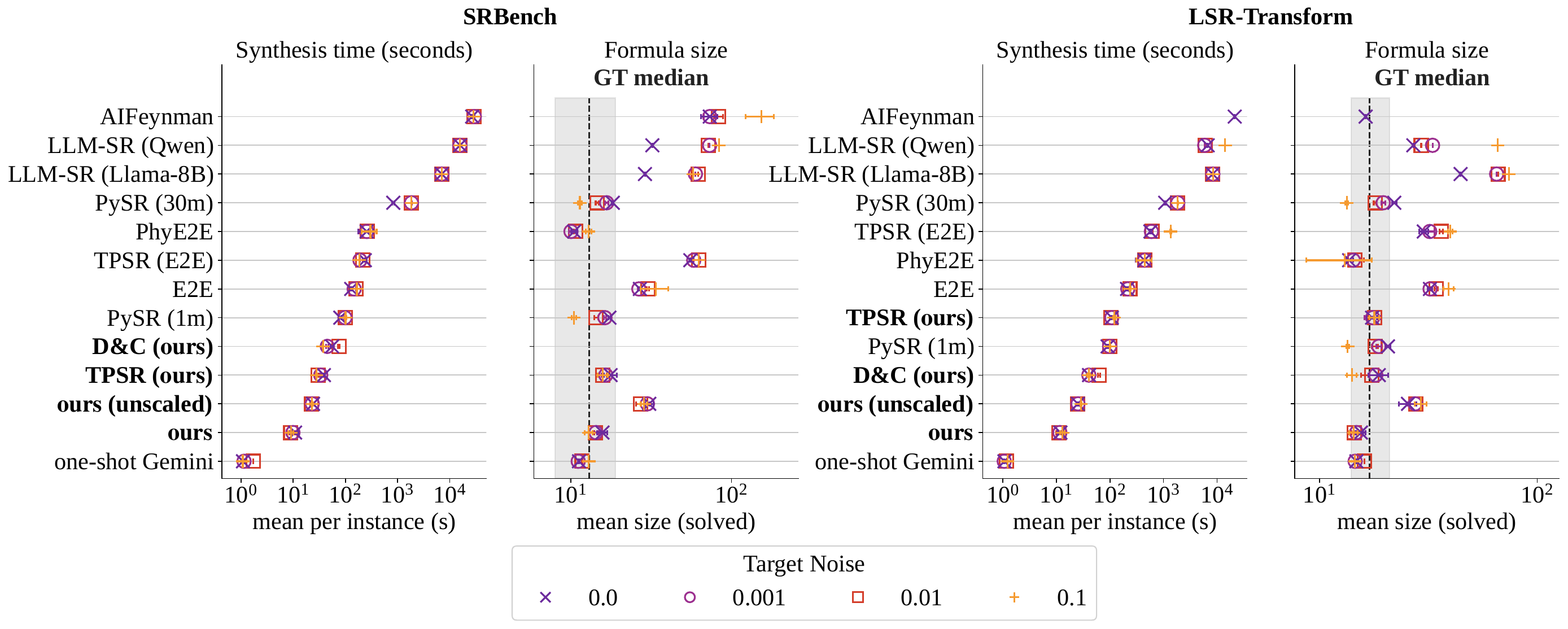}
\caption{Average per formula synthesis time and average synthesized formula size on the $ 99 $ SRBench formulae and on the $ 111 $ LSR-Transform formulae. The median formula size and standard deviation of the ground-truth formulae are shown by the dashed vertical line and the gray region.} 
\label{fig:both_time_complexity}
\end{figure*}

On SRBench the highest $ \rho_0 $ belongs to TPSR with the E2E model ($ 0.97 $), but its $ \rho_k $ falls to $ 0.38 $ at $ k = 128 $, whereas our transformer keeps $ \rho_{128} = 0.57 $, and $ 0.60 $ with D\&C (Figure~\ref{fig:both_extrapolation}). Without noise three methods reach a higher $ \rho_{128} $ than our transformer with or without D\&C but each requires more synthesis time (Figure~\ref{fig:both_time_complexity}): PySR (1m) reaches $ \rho_{128} = 0.62 $ in $ 1.3 $ minutes per formula against our $ 11 $ seconds, AI Feynman $ 0.69 $ in $ 7.5 $ hours, and PySR (30m) $ 0.78 $ in $ 14 $ minutes. Under target noise AI Feynman's performance degrades ($ \rho_{128} = 0.33 $ at $ \epsilon = 0.01 $) while our transformer's does not ($ 0.58 $, against $ 0.57 $ for PySR (1m) and $ 0.67 $ for PySR (30m)). LLM-SR with Llama-8B is comparable to our transformer without noise ($ 0.54 $) at about $ 2 $ hours per task, but degrades under noise. Applying TPSR to our transformer makes $ \rho_k $ worse than plain beam search even at $ k = 0 $, which we suspect is because its reward favors a close fit to $ D_\text{train} $ over a compact formula, while applying D\&C improves it only marginally, since it replaces the beam-search formula only on the tasks where that formula fails the held-out threshold $ \theta $.

At target noise level of $ 0.1 $ the highest $ \rho_0 $ on SRBench is again TPSR with the E2E model ($ 0.90 $), closely followed by {\em ours (unscaled)} (Section~\ref{subsec:inferenceMethods}) ($ 0.89 $) and LLM-SR with Llama-8B ($ 0.84 $), but all three fall below $ 0.05 $ by $ k = 128 $ (Figure~\ref{fig:both_extrapolation}). For every $ k \ge 2 $ PySR (30m) has the highest $ \rho_k $ ($ 0.51 $ at $ k = 128 $), followed by PySR (1m) ($ 0.39 $), against $ 0.26 $ for our transformer. One-shot Gemini has a substantially higher solution rate than the two one-shot open-source models (Figure~\ref{fig:appx_both_llm}) and is relatively robust to target noise.

On LSR-Transform (Figure~\ref{fig:both_extrapolation}) PySR (30m) has the highest $ \rho_k $ at every $ k $ up to target noise $ 0.01 $. Among the other methods TPSR with the E2E model has the highest $ \rho_0 $ ($ 0.73 $) but falls to $ 0.10 $ at $ k = 128 $. Up to target noise $ 0.01 $ our transformer, with or without D\&C, has the highest $ \rho_k $ for every $ k \ge 1 $ among the methods in Figure~\ref{fig:both_extrapolation} other than PySR (30m): $ \rho_{128} = 0.45 $ without noise and $ 0.39 $ at $ \epsilon = 0.01 $, against $ 0.33 $ and $ 0.31 $ for PySR (1m). PySR (30m) extrapolates better ($ 0.54 $ and $ 0.53 $) at about $ 18 $ minutes per task against our $ 12 $ seconds (Figure~\ref{fig:both_time_complexity}). At target noise $ 0.1 $ LLM-SR with Qwen has the highest $ \rho_0 $ ($ 0.71 $), and at $ k = 128 $ PySR (30m) reaches $ 0.35 $, PySR (1m) $ 0.15 $ and our transformer $ 0.12 $.

Due to API cost we ran LLM-SR with Gemini on LSR-Transform for a single run in the noise-free condition and compare it against the seed of each other method with the highest $ \rho_0 $ (Figure~\ref{fig:llmsr_extrapolation_time_complexity} in Appendix~\ref{app:bestseed}). There PySR (30m) has the highest $ \rho_0 $ ($ 0.88 $) and $ \rho_1 $, while LLM-SR with Gemini has the highest $ \rho_k $ for every $ k \ge 2 $ ($ 0.61 $ at $ k = 128 $, against $ 0.50 $ for PySR (30m) and $ 0.47 $ for our transformer), at about $ 2.2 $ hours per formula against about $ 18 $ minutes for PySR (30m) and $ 12 $ seconds for ours.

\begin{figure*}[t]
\centering
\includegraphics[width=0.8\textwidth]{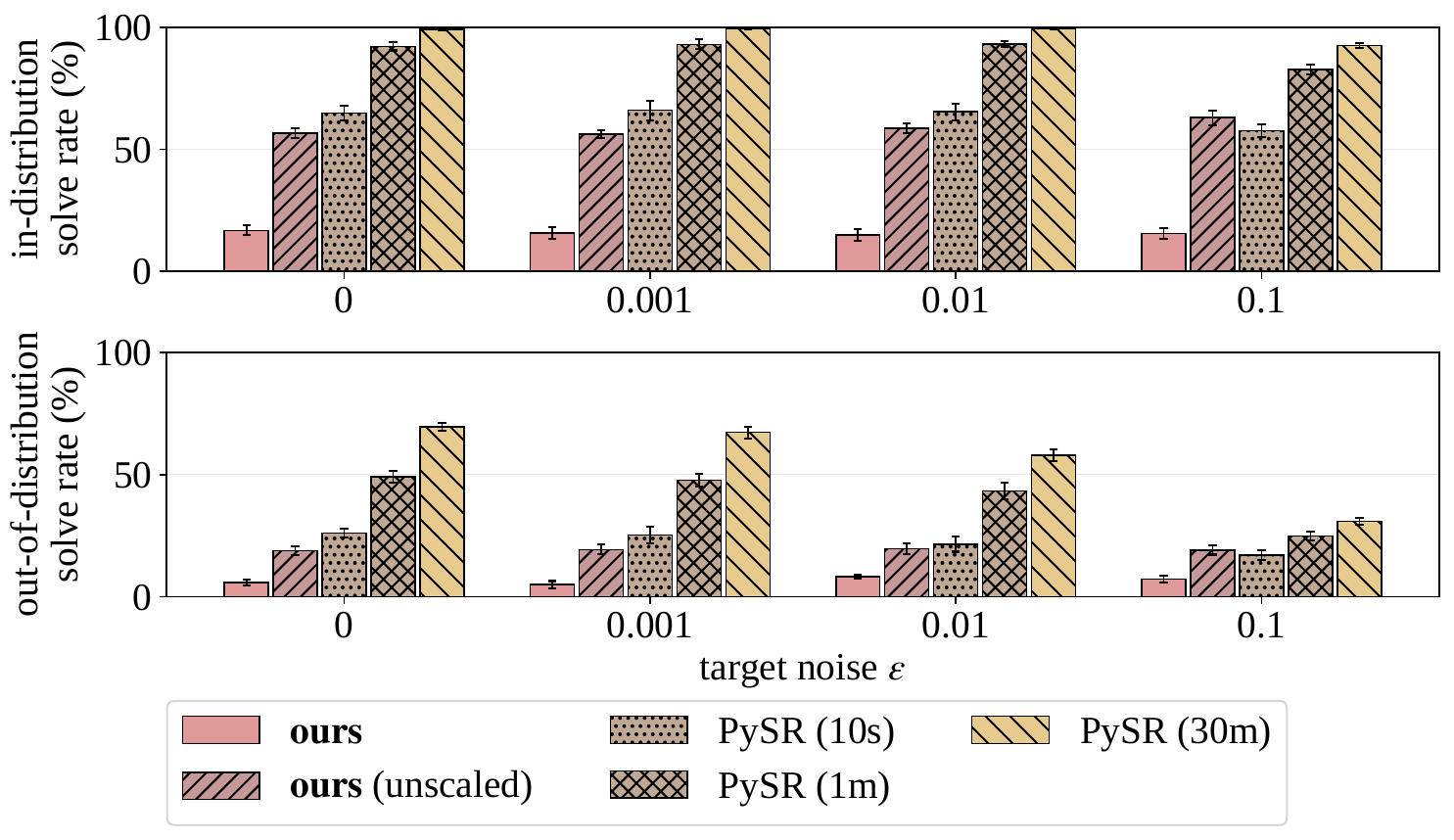}
\caption{In-distribution and out-of-distribution solution rate on the $ 129 $ LSR-Synth formulae.}
\label{fig:lsr_synth_noise}
\end{figure*}

For LSR-Synth we report the solution rate on the in-distribution and out-of-distribution test splits the benchmark provides (Figure~\ref{fig:lsr_synth_noise}). Without noise our transformer solves $ 17\% $ of the tasks in distribution and $ 6\% $ out of distribution, far below PySR (10s) ($ 65\% $ and $ 26\% $), PySR (1m) ($ 92\% $ and $ 49\% $) and PySR (30m) ($ 99\% $ and $ 70\% $). We attribute the gap to input range. Our raw training inputs lie in a narrow range around zero (Section~\ref{subsec:dataGeneration}), and those of SRBench and LSR-Transform lie mostly in $ [1, 5] $. LSR-Synth inputs instead have physical ranges, such as temperatures in $ [273, 543] $ in the material science domain. Reading the synthesized formula as a function of the whitened inputs, {\em ours (unscaled)}, brings them to our training range and raises our solution rate to $ 57\% $ in distribution and $ 19\% $ out of distribution. The largest gain is in the material science domain ($ 26\% $ to $ 84\% $), whose inputs are furthest from our training range, and the smallest in the physics domain ($ 9\% $ to $ 33\% $), whose inputs in $ [-1.3, 1.3] $ are already close to it. Search-based methods nonetheless remain ahead on LSR-Synth. We suspect two reasons: they fit the raw data directly and so need no assumption about range, and they refine candidates against the data over many iterations, whereas our transformer must name the whole formula in one pass.

\subsection{Ablation Study}
\label{subsec:ablation_study}

Our transformer synthesizes a formula in about $ 11 $ seconds and reaches $ \rho_{128} = 0.57 $ on SRBench, a higher solution rate out of distribution than any other method we evaluate that takes under a minute per task. This raises the question of which of our data-generation choices produce this gain, and whether either of them costs accuracy in distribution. We examine the two choices: dropping the affine transformation that places constants at every additive position (Section~\ref{subsec:dataGeneration}), and simplifying each formula before training.

\begin{figure*}[!htp]
\centering
\includegraphics[width=0.9\textwidth]{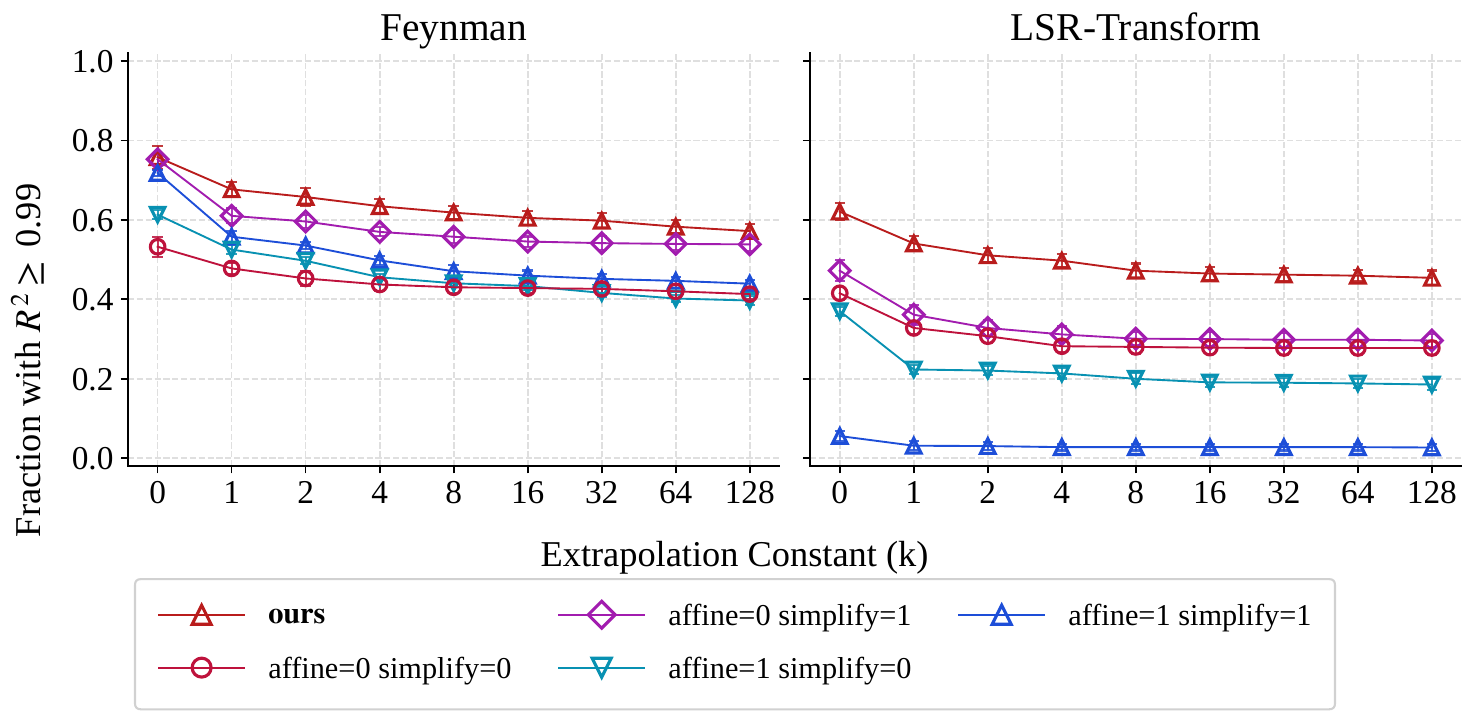}
\caption{The ablation study results on affine transformation and simplification.}
\label{fig:ablation_study}
\end{figure*}

We study the contribution of the affine transformation and simplification within our approach by training four transformers, one for each combination of the two components being present or absent from our approach. Each of them uses the $ 89 $ million-parameter architecture of~\citet{kamienny2022e2e} and is trained from scratch on $ 120 $ million generated training examples. We evaluate the checkpoint with the lowest validation loss on SRBench and LSR-Transform with $ 10 $ random seeds. On both benchmarks the transformer trained with simplification but without the affine transformation has the highest $ \rho_k $ (Figure~\ref{fig:ablation_study}).

%%%%%%%%%%%%%%%%%%%%%%%%%%%%%%%%%%%%%%%%%%%%%%%%%%%%%%%%%%%%

\section{Future Work}
\label{sec:future_work}

Our transformer supports up to ten variables, while a symbolic regression task can involve hundreds, not all of them relevant. Future work will explore transformers accepting more variables and methods to identify the relevant ones. Our raw training inputs lie in a narrow range around zero, unlike the physical ranges of LSR-Synth inputs, such as temperatures in $ [273, 543] $. Future work will draw the training input ranges at random, over several orders of magnitude, so that the transformer handles such inputs without the substitution of {\em ours (unscaled)}.

%%%%%%%%%%%%%%%%%%%%%%%%%%%%%%%%%%%%%%%%%%%%%%%%%%%%%%%%%%%%

\section{Conclusion}
\label{sec:conclusion}

We proposed a way of shaping the training data of a transformer for symbolic regression on physics formulae: each training formula is simplified into a compact, consistently ordered form. On SRBench and LSR-Transform the resulting transformer synthesizes a formula in about ten seconds and extrapolates better than every other method evaluated that takes under a minute per formula. We attribute this performance to the fact that our training formulae possibly have a structure more similar to the actual physics formulae than used in previous studies. Search-based methods can extrapolate better but only when given one to three orders of magnitude more synthesis time. Additional training on noise-corrupted targets keeps our solution rate $ \rho_k $ largely unchanged up to target noise $ 0.01 $, where only PySR with a $ 30 $-minute or longer budget extrapolates better on both benchmarks.

\section*{AI use statement}

We used AI for code implementation, for drafting and condensing parts of the text, for checking typos and grammatical errors, for creating tables in \LaTeX{} in the appendix, and for identifying related research. We have reviewed all AI contributions before inclusion in this paper and take responsibility for the full content of this paper.

\section*{Reproducibility statement}

Our source code is publicly available at \url{https://github.com/ShwayW/shorter-formulae}. It contains the formula generator and simplification method, the training script, one evaluation entry point per method, and the scripts that construct the extrapolation sets. It also includes the script that downloads the LLM-SRBench data and the configuration files that define each LLM method. The weights of our six transformers (the transformer we evaluate, the same transformer before fine-tuning on noisy targets, and the four transformers of Section~\ref{subsec:ablation_study}) are hosted on Hugging Face at \url{https://huggingface.co/ShwayW/shorter-formulae}. The other methods are run from their authors' released implementations and, where available, their released weights, with the settings given in Section~\ref{sec:empiricalEvaluation} and the Appendix.

\section*{Acknowledgments}
We thank Shivam Garg (University of Alberta) for helpful discussions. This work is supported by the Natural Sciences and Engineering Research Council of Canada (NSERC). Computational resources were provided by the Digital Research Alliance of Canada (DRAC).

\bibliographystyle{unsrtnat}
\bibliography{refs}  %%% Uncomment this line and comment out the ``thebibliography'' section below to use the external .bib file (using bibtex) .

\begin{thebibliography}{47}
\providecommand{\natexlab}[1]{#1}
\providecommand{\url}[1]{\texttt{#1}}
\expandafter\ifx\csname urlstyle\endcsname\relax
  \providecommand{\doi}[1]{doi: #1}\else
  \providecommand{\doi}{doi: \begingroup \urlstyle{rm}\Url}\fi

\bibitem[Schmidt and Lipson(2009)]{schmidt2009distilling}
Michael Schmidt and Hod Lipson.
\newblock Distilling free-form natural laws from experimental data.
\newblock \emph{Science}, 324:\penalty0 81--85, 2009.

\bibitem[Lample and Charton(2020)]{lample2020deep}
Guillaume Lample and Fran{\c{c}}ois Charton.
\newblock Deep learning for symbolic mathematics.
\newblock In \emph{Proceedings of International Conference on Learning
  Representations (ICLR)}, 2020.

\bibitem[Udrescu and Tegmark(2020)]{udrescu2020aifeynman}
Silviu-Marian Udrescu and Max Tegmark.
\newblock {AI Feynman}: A physics-inspired method for symbolic regression.
\newblock \emph{Science Advances}, 6\penalty0 (16):\penalty0 eaay2631, 2020.

\bibitem[La~Cava et~al.(2021)La~Cava, Orzechowski, Burlacu, de~Fran{\c{c}}a,
  Virgolin, Jin, Kommenda, and Moore]{lacava2021contemporary}
William La~Cava, Patryk Orzechowski, Bogdan Burlacu, Fabr{\'\i}cio~Olivetti
  de~Fran{\c{c}}a, Marco Virgolin, Ying Jin, Michael Kommenda, and Jason~H.
  Moore.
\newblock Contemporary symbolic regression methods and their relative
  performance.
\newblock In \emph{Proceedings of the Neural Information Processing Systems
  (NIPS) Track on Datasets and Benchmarks}, volume~1, 2021.

\bibitem[Biggio et~al.(2021)Biggio, Bendinelli, Neitz, Lucchi, and
  Parascandolo]{biggio2021nesymres}
Luca Biggio, Tommaso Bendinelli, Alexander Neitz, Aur{\'e}lien Lucchi, and
  Giambattista Parascandolo.
\newblock Neural symbolic regression that scales.
\newblock In \emph{Proceedings of International Conference on Machine Learning
  (ICML)}, pages 936--945, 2021.

\bibitem[Kamienny et~al.(2022)Kamienny, d'Ascoli, Lample, and
  Charton]{kamienny2022e2e}
Pierre-Alexandre Kamienny, St{\'e}phane d'Ascoli, Guillaume Lample, and
  Fran{\c{c}}ois Charton.
\newblock End-to-end symbolic regression with transformers.
\newblock In \emph{Proceedings of Advances in Neural Information Processing
  Systems (NeurIPS)}, volume~35, pages 10269--10281, 2022.

\bibitem[Shojaee et~al.(2023)Shojaee, Meidani, Barati~Farimani, and
  Reddy]{shojaee2023tpsr}
Parshin Shojaee, Kazem Meidani, Amir Barati~Farimani, and Chandan~K. Reddy.
\newblock Transformer-based planning for symbolic regression.
\newblock In \emph{Proceedings of Advances in Neural Information Processing
  Systems (NeurIPS)}, 2023.

\bibitem[Kubal{\'\i}k et~al.(2021)Kubal{\'\i}k, Derner, Zegklitz, and
  Babu{\v{s}}ka]{kubalik2021symbolic}
Ji{\v{r}}{\'\i} Kubal{\'\i}k, Erik Derner, Jan Zegklitz, and Robert
  Babu{\v{s}}ka.
\newblock Symbolic regression methods for reinforcement learning.
\newblock \emph{Institute of Electrical and Electronics Engineers (IEEE)
  Access}, 9:\penalty0 139697--139711, 2021.

\bibitem[Derner et~al.(2020)Derner, Kubal{\'\i}k, Ancona, and
  Babu{\v{s}}ka]{derner2020constructing}
Erik Derner, Ji{\v{r}}{\'\i} Kubal{\'\i}k, Nicola Ancona, and Robert
  Babu{\v{s}}ka.
\newblock Constructing parsimonious analytic models for dynamic systems via
  symbolic regression.
\newblock \emph{Applied Soft Computing}, 94:\penalty0 106432, 2020.

\bibitem[Voigt et~al.(2025)Voigt, Kahlmeyer, Lawonn, Habeck, and
  Giesen]{voigt2025analyzing}
Henrik Voigt, Paul Kahlmeyer, Kai Lawonn, Michael Habeck, and Joachim Giesen.
\newblock Analyzing generalization in pre-trained symbolic regression.
\newblock \emph{arXiv preprint arXiv:2509.19849}, 2025.

\bibitem[Schmidhuber(2009)]{schmidhuber2008driven}
J{\"u}rgen Schmidhuber.
\newblock Driven by compression progress: A simple principle explains essential
  aspects of subjective beauty, novelty, surprise, interestingness, attention,
  curiosity, creativity, art, science, music, jokes.
\newblock In \emph{Proceedings of the Workshop on Anticipatory Behavior in
  Adaptive Learning Systems (ABiALS)}, volume 5499, pages 48--76, 2009.

\bibitem[Segu{\'i} and Armour(2026)]{segui2026geometric}
Aleix Segu{\'i} and Wesley Armour.
\newblock A geometric lens on physics-aligned data compression.
\newblock In \emph{Proceedings of International Conference on Machine Learning
  (ICML)}, volume 306, 2026.

\bibitem[Schulte(2009)]{schulte2009simultaneous}
Oliver Schulte.
\newblock Simultaneous discovery of conservation laws and hidden particles with
  smith matrix decomposition.
\newblock In \emph{Proceedings of International Joint Conference on Artificial
  Intelligence (IJCAI)}, pages 1481--1487, 2009.

\bibitem[Shojaee et~al.(2025{\natexlab{a}})Shojaee, Nguyen, Meidani,
  Barati~Farimani, Doan, and Reddy]{shojaee2025llmsrbench}
Parshin Shojaee, Ngoc-Hieu Nguyen, Kazem Meidani, Amir Barati~Farimani, Khoa~D.
  Doan, and Chandan~K. Reddy.
\newblock {LLM-SRBench}: A new benchmark for scientific equation discovery with
  large language models.
\newblock In \emph{Proceedings of International Conference on Machine Learning
  (ICML)}, 2025{\natexlab{a}}.

\bibitem[Cranmer et~al.(2020)Cranmer, Sanchez~Gonzalez, Battaglia, Xu, Cranmer,
  Spergel, and Ho]{cranmer2020discovering}
Miles Cranmer, Alvaro Sanchez~Gonzalez, Peter Battaglia, Rui Xu, Kyle Cranmer,
  David Spergel, and Shirley Ho.
\newblock Discovering symbolic models from deep learning with inductive biases.
\newblock In \emph{Proceedings of Advances in Neural Information Processing
  Systems (NeurIPS)}, volume~33, pages 17429--17442, 2020.

\bibitem[Petersen et~al.(2021)Petersen, Larma, Mundhenk, Santiago, Kim, and
  Kim]{petersen2021deep}
Brenden~K. Petersen, Mikel~Landajuela Larma, Terrell~N. Mundhenk, Claudio~Prata
  Santiago, Soo~Kyung Kim, and Joanne~Taery Kim.
\newblock Deep symbolic regression: Recovering mathematical expressions from
  data via risk-seeking policy gradients.
\newblock In \emph{Proceedings of International Conference on Learning
  Representations (ICLR)}, 2021.

\bibitem[Bulitko et~al.(2022)Bulitko, Wang, Stevens, and
  Lelis]{bulitko2022portability}
Vadim Bulitko, Shuwei Wang, Justin Stevens, and Levi~HS Lelis.
\newblock Portability and explainability of synthesized formula-based
  heuristics.
\newblock In \emph{Proceedings of the International Symposium on Combinatorial
  Search (SoCS)}, pages 29--37, 2022.

\bibitem[Wang et~al.(2023)Wang, Bulitko, Huang, Koenig, and
  Stern]{wang2023synthesizing}
Shuwei Wang, Vadim Bulitko, Taoan Huang, Sven Koenig, and Roni Stern.
\newblock Synthesizing priority planning formulae for multi-agent pathfinding.
\newblock In \emph{Proceedings of the AAAI Conference on Artificial
  Intelligence and Interactive Digital Entertainment (AIIDE)}, pages 360--369,
  2023.

\bibitem[Sun et~al.(2023)Sun, Liu, Wang, and Sun]{sun2023symbolic}
Fangzheng Sun, Yang Liu, Jian-Xun Wang, and Hao Sun.
\newblock Symbolic physics learner: Discovering governing equations via monte
  carlo tree search.
\newblock In \emph{Proceedings of International Conference on Learning
  Representations (ICLR)}, 2023.

\bibitem[Cranmer(2023)]{cranmer2023interpretable}
Miles Cranmer.
\newblock Interpretable machine learning for science with pysr and
  symbolicregression. jl.
\newblock \emph{arXiv preprint arXiv:2305.01582}, 2023.

\bibitem[Grayeli et~al.(2024)Grayeli, Sehgal, Costilla-Reyes, Cranmer, and
  Chaudhuri]{grayeli2024symbolic}
Arya Grayeli, Atharva Sehgal, Omar Costilla-Reyes, Miles Cranmer, and Swarat
  Chaudhuri.
\newblock Symbolic regression with a learned concept library.
\newblock In \emph{Proceedings of Advances in Neural Information Processing
  Systems (NeurIPS)}, volume~37, pages 44678--44709, 2024.

\bibitem[Ma et~al.(2024)Ma, Wang, Guo, Sun, Tenenbaum, Rus, Gan, and
  Matusik]{ma2024llm}
Pingchuan Ma, Tsun-Hsuan Wang, Minghao Guo, Zhiqing Sun, Joshua~B. Tenenbaum,
  Daniela Rus, Chuang Gan, and Wojciech Matusik.
\newblock {LLM} and simulation as bilevel optimizers: A new paradigm to advance
  physical scientific discovery.
\newblock In \emph{Proceedings of International Conference on Machine Learning
  (ICML)}, 2024.

\bibitem[Hua et~al.(2025)Hua, Li, Yao, Zhuang, Tang, Liu, Ouyang, and
  Lu]{hua2025finetuning}
Yingfan Hua, Ruikun Li, Jun Yao, Guohang Zhuang, Shixiang Tang, Bin Liu, Wanli
  Ouyang, and Yan Lu.
\newblock Finetuning large language model as an effective symbolic regressor.
\newblock \emph{arXiv preprint arXiv:2508.09897}, 2025.

\bibitem[Shojaee et~al.(2025{\natexlab{b}})Shojaee, Meidani, Gupta,
  Barati~Farimani, and Reddy]{shojaee2025llmsr}
Parshin Shojaee, Kazem Meidani, Shashank Gupta, Amir Barati~Farimani, and
  Chandan~K. Reddy.
\newblock {LLM-SR}: Scientific equation discovery via programming with large
  language models.
\newblock In \emph{Proceedings of International Conference on Learning
  Representations (ICLR)}, 2025{\natexlab{b}}.

\bibitem[Valipour et~al.(2022)Valipour, You, Panju, and
  Ghodsi]{valipour2022symbolicgpt}
Mojtaba Valipour, Bowen You, Maysum~H. Panju, and Ali Ghodsi.
\newblock {SymbolicGPT}: A generative transformer model for symbolic
  regression.
\newblock In \emph{Proceedings of the Workshop on Efficient Natural Language
  and Speech Processing (ENLSP)}, 2022.

\bibitem[d'Ascoli et~al.(2024)d'Ascoli, Becker, Schwaller, Mathis, and
  Kilbertus]{d2024odeformer}
St{\'e}phane d'Ascoli, S{\"o}ren Becker, Philippe Schwaller, Alexander Mathis,
  and Niki Kilbertus.
\newblock {ODEFormer}: Symbolic regression of dynamical systems with
  transformers.
\newblock In \emph{Proceedings of International Conference on Learning
  Representations (ICLR)}, volume 2024, pages 21943--21976, 2024.

\bibitem[Vastl et~al.(2024)Vastl, Kulh{\'a}nek, Kubal{\'\i}k, Derner, and
  Babu{\v{s}}ka]{vastl2024symformer}
Martin Vastl, Jon{\'a}{\v{s}} Kulh{\'a}nek, Ji{\v{r}}{\'\i} Kubal{\'\i}k, Erik
  Derner, and Robert Babu{\v{s}}ka.
\newblock Symformer: End-to-end symbolic regression using transformer-based
  architecture.
\newblock \emph{IEEE Access}, 12:\penalty0 37840--37849, 2024.

\bibitem[Ying et~al.(2025)Ying, Lin, Yue, Chen, Xiao, Shi, Liang, Yau, Zhou,
  and Ma]{ying2025neural}
Jie Ying, Haowei Lin, Chao Yue, Yajie Chen, Chao Xiao, Quanqi Shi, Yitao Liang,
  Shing-Tung Yau, Yuan Zhou, and Jianzhu Ma.
\newblock A neural symbolic model for space physics.
\newblock \emph{Nature Machine Intelligence}, 7:\penalty0 1726--1741, 2025.

\bibitem[Bendinelli et~al.(2023)Bendinelli, Biggio, and
  Kamienny]{bendinelli2023controllable}
Tommaso Bendinelli, Luca Biggio, and Pierre-Alexandre Kamienny.
\newblock Controllable neural symbolic regression.
\newblock In \emph{Proceedings of International Conference on Machine Learning
  (ICML)}, pages 2063--2077, 2023.

\bibitem[Saegert and K{\"o}the(2026)]{saegert2026simplipy}
Paul Saegert and Ullrich K{\"o}the.
\newblock Breaking the simplification bottleneck in amortized neural symbolic
  regression.
\newblock In \emph{Proceedings of International Conference on Machine Learning
  (ICML)}, volume 306, 2026.

\bibitem[Meurer et~al.(2017)Meurer, Smith, Paprocki, {\v{C}}ert{\'\i}k,
  Kirpichev, Rocklin, Kumar, Ivanov, Moore, Singh, et~al.]{meurer2017sympy}
Aaron Meurer, Christopher~P Smith, Mateusz Paprocki, Ond{\v{r}}ej
  {\v{C}}ert{\'\i}k, Sergey~B Kirpichev, Matthew Rocklin, AMiT Kumar, Sergiu
  Ivanov, Jason~K Moore, Sartaj Singh, et~al.
\newblock Sympy: symbolic computing in python.
\newblock \emph{PeerJ Computer Science}, 3:\penalty0 e103, 2017.

\bibitem[Udrescu et~al.(2020)Udrescu, Tan, Feng, Neto, Wu, and
  Tegmark]{udrescu2020ai}
Silviu-Marian Udrescu, Andrew Tan, Jiahai Feng, Orisvaldo Neto, Tailin Wu, and
  Max Tegmark.
\newblock {AI Feynman 2.0}: Pareto-optimal symbolic regression exploiting graph
  modularity.
\newblock In \emph{Proceedings of Advances in Neural Information Processing
  Systems (NeurIPS)}, volume~33, pages 4860--4871, 2020.

\bibitem[Matsubara et~al.(2024)Matsubara, Chiba, Igarashi, and
  Ushiku]{matsubara2024rethinking}
Yoshitomo Matsubara, Naoya Chiba, Ryo Igarashi, and Yoshitaka Ushiku.
\newblock Rethinking symbolic regression datasets and benchmarks for scientific
  discovery.
\newblock \emph{Journal of Data-centric Machine Learning Research (DMLR)},
  2024.

\bibitem[Richardson(1968)]{richardson1968}
Daniel Richardson.
\newblock Some undecidable problems involving elementary functions of a real
  variable.
\newblock \emph{Journal of Symbolic Logic}, 33\penalty0 (4):\penalty0 514--520,
  1968.

\bibitem[Cohen(2003)]{cohen2003computer}
Joel~S Cohen.
\newblock \emph{Computer algebra and symbolic computation: Mathematical
  methods}.
\newblock CRC Press, 2003.

\bibitem[Charton(2022)]{charton2022linear}
Fran{\c{c}}ois Charton.
\newblock Linear algebra with transformers.
\newblock \emph{Transactions on Machine Learning Research (TMLR)}, 2022.

\bibitem[Vaswani et~al.(2017)Vaswani, Shazeer, Parmar, Uszkoreit, Jones, Gomez,
  Kaiser, and Polosukhin]{vaswani2017attention}
Ashish Vaswani, Noam Shazeer, Niki Parmar, Jakob Uszkoreit, Llion Jones,
  Aidan~N Gomez, {\L}ukasz Kaiser, and Illia Polosukhin.
\newblock Attention is all you need.
\newblock In \emph{Proceedings of Advances in Neural Information Processing
  Systems (NeurIPS)}, volume~30, 2017.

\bibitem[Phuong and Hutter(2022)]{phuong2022formal}
Mary Phuong and Marcus Hutter.
\newblock Formal algorithms for transformers.
\newblock \emph{arXiv preprint arXiv:2207.09238}, 2022.

\bibitem[Xiong et~al.(2020)Xiong, Yang, He, Zheng, Zheng, Xing, Zhang, Lan,
  Wang, and Liu]{xiong2020layer}
Ruibin Xiong, Yunchang Yang, Di~He, Kai Zheng, Shuxin Zheng, Chen Xing,
  Huishuai Zhang, Yanyan Lan, Liwei Wang, and Tieyan Liu.
\newblock On layer normalization in the transformer architecture.
\newblock In \emph{Proceedings of International Conference on Machine Learning
  (ICML)}, pages 10524--10533, 2020.

\bibitem[Hendrycks and Gimpel(2016)]{hendrycks2016gaussian}
Dan Hendrycks and Kevin Gimpel.
\newblock Gaussian error linear units ({GELUs}).
\newblock \emph{arXiv preprint arXiv:1606.08415}, 2016.

\bibitem[Ba and Kingma(2015)]{kingma2015adam}
Jimmy Ba and Diederik~P. Kingma.
\newblock Adam: A method for stochastic optimization.
\newblock In \emph{Proceedings of International Conference on Learning
  Representations (ICLR)}, 2015.

\bibitem[Wiseman and Rush(2016)]{wiseman2016sequence}
Sam Wiseman and Alexander~M Rush.
\newblock Sequence-to-sequence learning as beam-search optimization.
\newblock In \emph{Proceedings of Conference on Empirical Methods in Natural
  Language Processing (EMNLP)}, pages 1296--1306, 2016.

\bibitem[Fletcher(2013)]{fletcher2013practical}
Roger Fletcher.
\newblock \emph{Practical methods of optimization}.
\newblock John Wiley \& Sons, 2013.

\bibitem[Lalande et~al.(2023)Lalande, Matsubara, Chiba, Taniai, Igarashi, and
  Ushiku]{lalande2023transformer}
Florian Lalande, Yoshitomo Matsubara, Naoya Chiba, Tatsunori Taniai, Ryo
  Igarashi, and Yoshitaka Ushiku.
\newblock A transformer model for symbolic regression towards scientific
  discovery.
\newblock \emph{arXiv preprint arXiv:2312.04070}, 2023.

\bibitem[Grattafiori et~al.(2024)Grattafiori, Dubey, Jauhri, Pandey, Kadian,
  Al-Dahle, Letman, Mathur, Schelten, Vaughan, et~al.]{grattafiori2024llama}
Aaron Grattafiori, Abhimanyu Dubey, Abhinav Jauhri, Abhinav Pandey, Abhishek
  Kadian, Ahmad Al-Dahle, Aiesha Letman, Akhil Mathur, Alan Schelten, Alex
  Vaughan, et~al.
\newblock The llama 3 herd of models.
\newblock \emph{arXiv preprint arXiv:2407.21783}, 2024.

\bibitem[Hui et~al.(2024)Hui, Yang, Cui, Yang, Liu, Zhang, Liu, Zhang, Yu, Lu,
  et~al.]{hui2024qwen2}
Binyuan Hui, Jian Yang, Zeyu Cui, Jiaxi Yang, Dayiheng Liu, Lei Zhang, Tianyu
  Liu, Jiajun Zhang, Bowen Yu, Keming Lu, et~al.
\newblock Qwen2. 5-coder technical report.
\newblock \emph{arXiv preprint arXiv:2409.12186}, 2024.

\bibitem[Lee et~al.(2019)Lee, Lee, Kim, Kosiorek, Choi, and Teh]{lee2019set}
Juho Lee, Yoonho Lee, Jungtaek Kim, Adam Kosiorek, Seungjin Choi, and Yee~Whye
  Teh.
\newblock Set transformer: A framework for attention-based
  permutation-invariant neural networks.
\newblock In \emph{Proceedings of International Conference on Machine Learning
  (ICML)}, 2019.

\end{thebibliography}

\appendix
\section{Appendix}

\subsection{Operator sampling weights}
\label{app:operatorWeights}

Table~\ref{tab:operatorWeights} lists the unnormalized sampling weight of each operator. We set these weights by hand, before any evaluation, and did not tune them on the benchmarks. They encode a domain-general ordering: arithmetic first, then squares, roots and reciprocals, then the core transcendental operators, and finally the rarer inverse trigonometric ones. When labeling a binary node of the skeleton we draw a binary operator with probability equal to its weight divided by the sum of the binary weights ($ 3.5 $), and when labeling a unary node we draw a unary operator with probability equal to its weight divided by the sum of the unary weights ($ 6.35 $). For example, \texttt{+} is drawn with probability $ 1/3.5 \approx 0.286 $ at a binary node and \texttt{sqr} with probability $ 1/6.35 \approx 0.157 $ at a unary node.

\begin{table}[H]
\centering
\caption{Unnormalized sampling weight of each operator. The probability of drawing an operator is its weight divided by the sum of the weights in its group (binary or unary). \texttt{pow2} is identical to \texttt{sqr} and is kept only for compatibility with the token vocabulary.}
\begin{tabular}{llr@{\hskip 2em}llr}
\toprule
\multicolumn{3}{l}{\em Binary} & \multicolumn{3}{l}{\em Unary (continued)} \\
\midrule
$ S + S $        & \texttt{+}      & $ 1 $ & $ S-\!- $     & \texttt{-\,-}   & $ 0.35 $ \\
$ S \times S $   & \texttt{*}      & $ 1 $ & $ |S| $       & \texttt{abs}    & $ 0.2 $ \\
$ S - S $        & \texttt{-}      & $ 0.8 $ & $ S^3 $       & \texttt{pow3}   & $ 0.2 $ \\
$ S / S $        & \texttt{/}      & $ 0.6 $ & $ \tan(S) $   & \texttt{tan}    & $ 0.12 $ \\
$ S^S $          & \texttt{pow}    & $ 0.1 $ & $ \arcsin(S) $& \texttt{arcsin} & $ 0.1 $ \\
\cmidrule(r){1-3}
\multicolumn{3}{l}{\em Unary}                  & $ \arctan(S) $& \texttt{arctan} & $ 0.1 $ \\
\cmidrule(r){1-3}
$ S^2 $          & \texttt{sqr}    & $ 1 $ & $ \tanh(S) $  & \texttt{tanh}   & $ 0.1 $ \\
$ \sqrt{S} $     & \texttt{sqrt}   & $ 0.7 $ & $ \arccos(S) $& \texttt{arccos} & $ 0.08 $ \\
$ -S $           & \texttt{neg}    & $ 0.6 $ & $ S^2 $       & \texttt{pow2}   & $ 0.05 $ \\
$ 1/S $          & \texttt{invert} & $ 0.6 $ &               &                 &            \\
$ \sin(S) $      & \texttt{sin}    & $ 0.5 $ &               &                 &            \\
$ \cos(S) $      & \texttt{cos}    & $ 0.5 $ &               &                 &            \\
$ e^S $          & \texttt{exp}    & $ 0.4 $ &               &                 &            \\
$ \ln(S) $       & \texttt{ln}     & $ 0.4 $ &               &                 &            \\
$ S+\!+ $        & \texttt{++}     & $ 0.35 $ &               &                 &            \\
\bottomrule
\end{tabular}
\label{tab:operatorWeights}
\end{table}

\subsection{Simplification method}
\label{app:simplification}

The pseudocode for the simplification method is shown in Algorithm~\ref{alg:simplify}, with its subroutines in Algorithms~\ref{alg:collectsum}, \ref{alg:collectprod} and~\ref{alg:helpers}. The cancellations $ \ln e^S = S $, $ -(-S) = S $, $ ||S|| = |S| $, $ e^{\ln S} = S $ and $ 1/(1/S) = S $ do not change the value of a formula at any input at which the original is defined. Of these, $ \ln e^S = S $, $ e^{\ln S} = S $ and $ ||S|| = |S| $ are applied as rewrites when a unary operator is normalized (\textsc{NormFunc} of Algorithm~\ref{alg:simplify}), while $ -(-S) = S $ and $ 1/(1/S) = S $ need no rule of their own: the flattening negates every coefficient of a sum for each $ -S $ and every exponent of a product for each $ 1/S $, and doing either twice is the identity. Some of the cancellations make a formula defined at additional inputs: merging exponents can cancel a factor against its reciprocal, as in $ S \cdot \frac{1}{S} \to 1 $, and $ e^{\ln S} = S $ removes the requirement that $ S $ be positive. The inputs affected are those at which a cancelled subformula is zero or leaves the domain of a cancelled operator, and they do not affect the training data: we generate the data points from the simplified formula rather than from the original draw, and discard any input at which it does not evaluate to a finite value.

Sums and products are sorted to place variables first in order of their index and all other subformulae in lexicographical order, so $ v_{10} + v_2 $ becomes $ v_2 + v_{10} $, and every variable precedes every non-variable subformula, so $ \cos(v_1) + v_3 $ becomes $ v_3 + \cos(v_1) $. Sorting applies within groups: when a sum is written back out its positive terms come before its negative ones, and a product its coefficient and numerator before its denominator, so $ v_2 - v_1 $ and $ v_3 / v_1 $ are already in simplified form.

\begin{algorithm}[htbp]
\caption{\textsc{Simplify}}
\label{alg:simplify}
\begin{algorithmic}[1]
\Function{Simplify}{$s$} \Comment{$ s $ is a given formula in standard Polish notation}
  \State Parse $ s $ into a formula tree $ T = (\mathrm{op}, \mathrm{kids}) $
  \State $I \gets \Call{Norm}{T}$ \Comment{Apply sum-of-products normalization to form an internal representation $ I $}
  \State \Return standard Polish notation converted from the internal representation $ I $
\EndFunction

\Statex
\Function{Norm}{$T = (\mathrm{op}, \mathrm{kids})$}
  \If{$\mathrm{op}$ is a leaf} \Comment{zero-arity: nothing to collect}
    \State \Return $ v_i $ for a variable, or $ c $ for a numeric constant
  \ElsIf{$\mathrm{op}$ is additive} \Comment{$ + $, $ - $, $ ++ $, $ -- $, $ \mathrm{neg} $}
    \State $(c_0, \{c_i, t_i\}) \gets \Call{CollectSum}{T}$
    \State \Return \Call{MakeSum}{$c_0, \{t_i \mapsto c_i\}$} \Comment{drops zero coefficients and sorts the terms}
  \ElsIf{$\mathrm{op}$ is multiplicative} \Comment{$ \times $, $ / $, $ \mathrm{invert} $, $ \mathrm{sqr} $, $ \mathrm{pow2} $, $ \mathrm{pow3} $}
    \State $(c, \{b_i \mapsto e_i\}) \gets \Call{CollectProd}{T}$
    \State \Return \Call{MakeProd}{$c, \{b_i \mapsto e_i\}$} \Comment{drops zero exponents and sorts the factors}
  \ElsIf{$\mathrm{op}$ is $ \mathrm{pow} $}
    \State \Return \Call{NormPow}{$\mathrm{kids}[0], \mathrm{kids}[1]$}
  \Else \Comment{$ \sqrt{\cdot} $, $ |\cdot| $, and the transcendental unary operators}
	\State $ A \gets \Call{Norm}{\mathrm{kids}[0]} $
	\State \Return \Call{NormFunc}{$ \mathrm{op}, A $}
  \EndIf
\EndFunction

\Statex
\Function{NormPow}{$\mathrm{base}, \mathrm{expNode}$} \Comment{$ \mathrm{base}^{\mathrm{expNode}} $; only integer exponents are merged}
  \State $ E \gets \Call{Norm}{\mathrm{expNode}} $
  \If{$ E $ is a constant $ n $}
	\If{$ n = 0 $}
	  \State \Return $ 1 $
	\ElsIf{$ n \in \ZZ $}
	  \State $ (\mathrm{c}, \mathrm{f}) \gets \Call{CollectProd}{\mathrm{base}} $
	  \If{$ n < 0 $ and $ \mathrm{c} = 0 $}
		\State \textbf{abort} \Comment{$ 0^n $ with $ n < 0 $ is undefined; the formula is left unchanged}
	  \EndIf
	  \State \Return \Call{MakeProd}{$ \mathrm{c}^{n}, \{b \mapsto e \times n \mid (b, e) \in \mathrm{f}\} $}
	\ElsIf{$ n = 1/2 $}
	  \State $ A \gets \Call{Norm}{\mathrm{base}} $
	  \State \Return \Call{NormFunc}{$ \sqrt{\cdot}, A $}
	\EndIf
  \EndIf
  \State \Return the opaque node $ (\Call{Norm}{\mathrm{base}})^{E} $ \Comment{symbolic or non-integer exponent}
\EndFunction

\Statex
\Function{NormFunc}{$\mathrm{op}, A$} \Comment{$ A $ is the already normalized argument}
  \If{$ \mathrm{op} $ is $ e^{\cdot} $ and $ A $ is $ \ln(A') $}
      \State \Return $ A' $ \Comment{$ e^{\ln S} = S $}
  \ElsIf{$ \mathrm{op} $ is $ \ln $ and $ A $ is $ e^{A'} $}
      \State \Return $ A' $ \Comment{$ \ln e^S = S $}
  \ElsIf{$ \mathrm{op} $ is $ |\cdot| $}
      \If{$ A $ is a constant $ c $}
        \State \Return $ |c| $
      \ElsIf{$ A $ is $ |A'| $}
        \State \Return $ A $ \Comment{$ ||S|| = |S| $}
      \EndIf
  \EndIf
  \State \Return the opaque node $ \mathrm{op}(A) $
\EndFunction
\end{algorithmic}
\end{algorithm}

\begin{algorithm}[t]
\caption{\textsc{CollectSum}}
\label{alg:collectsum}
\begin{algorithmic}[1]
\Function{CollectSum}{$T = (\mathrm{op}, \mathrm{kids})$} \Comment{returns $(c_0, \{t_i \mapsto c_i\})$ with value $c_0 + \sum_i c_i t_i$}
  \If{$ \mathrm{op} $ is $ + $}
	\State $ (\mathrm{cL}, \mathrm{tL}) \gets \Call{CollectSum}{\mathrm{kids}[0]} $
	\State $ (\mathrm{cR}, \mathrm{tR}) \gets \Call{CollectSum}{\mathrm{kids}[1]} $
	\State \Return $ (\mathrm{cL} + \mathrm{cR}, \Call{Merge}{\mathrm{tL}, \mathrm{tR}}) $ \Comment{a shared base has its coefficients added}
  \ElsIf{$ \mathrm{op} $ is $ - $}
	\State $ (\mathrm{cL}, \mathrm{tL}) \gets \Call{CollectSum}{\mathrm{kids}[0]} $
	\State $ (\mathrm{cR}, \mathrm{tR}) \gets \Call{CollectSum}{\mathrm{kids}[1]} $
	\State $ \mathrm{tR}' \gets \Call{Neg}{\mathrm{tR}} $ \Comment{subtracting negates every coefficient}
	\State \Return $ (\mathrm{cL} - \mathrm{cR}, \Call{Merge}{\mathrm{tL}, \mathrm{tR}'}) $
  \ElsIf{$ \mathrm{op} $ is $ ++ $}
	\State $ (\mathrm{c}, \mathrm{t}) \gets \Call{CollectSum}{\mathrm{kids}[0]} $
	\State \Return $ (\mathrm{c} + 1, \mathrm{t}) $
  \ElsIf{$ \mathrm{op} $ is $ -- $}
	\State $ (\mathrm{c}, \mathrm{t}) \gets \Call{CollectSum}{\mathrm{kids}[0]} $
	\State \Return $ (\mathrm{c} - 1, \mathrm{t}) $
  \ElsIf{$ \mathrm{op} $ is $ \mathrm{neg} $}
	\State $ (\mathrm{c}, \mathrm{t}) \gets \Call{CollectSum}{\mathrm{kids}[0]} $
	\State \Return $ (-\mathrm{c}, \Call{Neg}{\mathrm{t}}) $
  \EndIf
  \State $ I \gets \Call{Norm}{T} $ \Comment{$ T $ is not additive: normalize it and contribute it as a single term}
  \If{$ I $ is a sum $ (c_0, \{(c_i, t_i)\}_i) $}
	\State \Return $ (c_0, \{c_i, t_i\}_i) $
  \EndIf
  \State $ (s, b) \gets \Call{SplitScalar}{I} $
  \If{$ b = \emptyset $}
	\State \Return $ (s, \emptyset) $ \Comment{$ I $ is a pure constant}
  \EndIf
  \State \Return $ (0, \{b \mapsto s\}) $
\EndFunction
\end{algorithmic}
\end{algorithm}

\begin{algorithm}[htbp]
\caption{\textsc{CollectProd}}
\label{alg:collectprod}
\begin{algorithmic}[1]
\Function{CollectProd}{$T = (\mathrm{op}, \mathrm{kids})$} \Comment{returns $(c, \{b_i \mapsto e_i\})$ with value $c \prod_i b_i^{e_i}$}
  \If{$ \mathrm{op} $ is $ \times $}
	\State $ (\mathrm{cL}, \mathrm{fL}) \gets \Call{CollectProd}{\mathrm{kids}[0]} $
	\State $ (\mathrm{cR}, \mathrm{fR}) \gets \Call{CollectProd}{\mathrm{kids}[1]} $
	\State \Return $ (\mathrm{cL} \times \mathrm{cR}, \Call{Merge}{\mathrm{fL}, \mathrm{fR}}) $ \Comment{a shared base has its exponents added}
  \ElsIf{$ \mathrm{op} $ is $ / $}
	\State $ (\mathrm{cL}, \mathrm{fL}) \gets \Call{CollectProd}{\mathrm{kids}[0]} $
	\State $ (\mathrm{cR}, \mathrm{fR}) \gets \Call{CollectProd}{\mathrm{kids}[1]} $
	\If{$ \mathrm{cR} = 0 $}
	  \State \textbf{abort} \Comment{division by zero is undefined; the formula is left unchanged}
	\EndIf
	\State $ \mathrm{fR}' \gets \Call{Neg}{\mathrm{fR}} $ \Comment{dividing subtracts the divisor's exponents}
	\State \Return $ (\mathrm{cL} / \mathrm{cR}, \Call{Merge}{\mathrm{fL}, \mathrm{fR}'}) $
  \ElsIf{$ \mathrm{op} $ is $ \mathrm{invert} $}
	\State $ (\mathrm{c}, \mathrm{f}) \gets \Call{CollectProd}{\mathrm{kids}[0]} $
	\If{$ \mathrm{c} = 0 $}
	  \State \textbf{abort} \Comment{$ 1/0 $ is undefined; the formula is left unchanged}
	\EndIf
	\State \Return $ (1 / \mathrm{c}, \Call{Neg}{\mathrm{f}}) $ \Comment{the coefficient is reciprocated, every exponent negated}
  \ElsIf{$ \mathrm{op} $ is in $ \{\mathrm{sqr}, \mathrm{pow2}\} $}
	\State $ (\mathrm{c}, \mathrm{f}) \gets \Call{CollectProd}{\mathrm{kids}[0]} $
	\State \Return $ (\mathrm{c}^2, \{b \mapsto e \times 2 \mid (b, e) \in \mathrm{f}\}) $
  \ElsIf{$ \mathrm{op} $ is $ \mathrm{pow3} $}
	\State $ (\mathrm{c}, \mathrm{f}) \gets \Call{CollectProd}{\mathrm{kids}[0]} $
	\State \Return $ (\mathrm{c}^3, \{b \mapsto e \times 3 \mid (b, e) \in \mathrm{f}\}) $
  \EndIf
  \State $ I \gets \Call{Norm}{T} $ \Comment{$ T $ is not multiplicative: normalize it and contribute it as a single factor}
  \State $ (s, b) \gets \Call{SplitScalar}{I} $
  \If{$ b = \emptyset $}
	\State \Return $ (s, \emptyset) $ \Comment{$ I $ is a pure constant}
  \ElsIf{$ b $ is a product $ (c', \mathrm{f}') $}
	\State \Return $ (s \times c', \mathrm{f}') $ \Comment{absorb the inner product}
  \EndIf
  \State \Return $ (s, \{b \mapsto 1\}) $
\EndFunction
\end{algorithmic}
\end{algorithm}

\begin{algorithm}[htbp]
\caption{\textsc{Helper routines}}
\label{alg:helpers}
\begin{algorithmic}[1]
\Function{Merge}{$m_1, m_2$} \Comment{union of the keys; the values are added on a shared key}
  \State \Return $ \{k \mapsto m_1[k] + m_2[k] \mid k \in \mathrm{keys}(m_1) \cup \mathrm{keys}(m_2)\} $ \Comment{an absent key contributes $ 0 $}
\EndFunction

\Statex
\Function{Neg}{$m$} \Comment{negate every value: subtracts terms in a sum, inverts factors in a product}
  \State \Return $ \{k \mapsto -m[k] \mid k \in \mathrm{keys}(m)\} $
\EndFunction

\Statex
\Function{SplitScalar}{$I$} \Comment{writes $ I = s \cdot b $ with a rational $ s $ and a monic base $ b $}
  \If{$ I $ is a rational constant $ c $}
	\State \Return $ (c, \emptyset) $ \Comment{$ I $ is a pure constant; there is no base}
  \ElsIf{$ I $ is a product $ (c, \mathrm{f}) $}
	\State \Return $ (c, \Call{MakeProd}{1, \mathrm{f}}) $ \Comment{strip the coefficient}
  \ElsIf{$ I $ is a sum $ (0, \{(c, b)\}) $}
	\State \Return $ (c, b) $ \Comment{a lone $ c \cdot b $ term}
  \EndIf
  \State \Return $ (1, I) $
\EndFunction
\end{algorithmic}
\end{algorithm}

\subsection{Oracle network configuration}
\label{app:oracleNetwork}

The divide-and-conquer strategy of Section~\ref{subsec:inferenceMethods} uses a feed-forward neural network as a differentiable proxy for the unknown function. It is queried for its second-order derivatives, which determine how the variables are grouped, and for the data points of each group once that grouping is fixed. Table~\ref{tab:oracleNetwork} gives its configuration.

\begin{table}[H]
\centering
\caption{Configuration of the oracle network used by the divide-and-conquer strategy.}
\begin{tabular}{ll}
\toprule
{\bf Parameter} & {\bf Value} \\
\midrule
\multicolumn{2}{l}{\em Architecture} \\
\quad hidden layers          & $ 128, 128, 64, 64 $ \\
\quad hidden activation      & $ \tanh $ \\
\quad output layer           & linear, one unit \\
\quad weights                & $ 128m + 29{,}121 $ for $ m $ variables \\
\cmidrule(r){1-2}
\multicolumn{2}{l}{\em Input and target scaling} \\
\quad inputs                 & whitened per variable \\
\quad target                 & standardized \\
\cmidrule(r){1-2}
\multicolumn{2}{l}{\em Training} \\
\quad loss                   & root mean squared error \\
\quad optimizer              & Adam, $ \beta = (0.9, 0.999) $ \\
\quad initial learning rate  & $ 1.0 \times 10^{-2} $ \\
\quad learning rate decay    & $ \div 10 $ after $ 20 $ epochs without improvement \\
\quad stopping criterion     & learning rate below $ 1.0 \times 10^{-5} $ \\
\quad epoch cap              & $ 3{,}000 $ \\
\quad batch size             & $ 256 $ \\
\cmidrule(r){1-2}
\multicolumn{2}{l}{\em Data} \\
\quad data points            & $ \le 20{,}000 $, drawn from $ D_\text{train} $ only \\
\quad validation split       & $ 20\% $ \\
\quad derivative sample      & $ 1{,}000 $ data points \\
\bottomrule
\end{tabular}
\label{tab:oracleNetwork}
\end{table}

Both separability tests of Section~\ref{subsec:inferenceMethods} are evaluated at $ 1{,}000 $ points drawn from $ D_\text{train} $ rather than from a box over the input range, since the network is only trustworthy where it saw data. Each pair of variables is scored by a dimensionless quantity so that a single threshold serves every task. For the additive test the score is the median of $ |H_{ij}| \, |f| \, / \, (|\partial_i f| \, |\partial_j f|) $ over those points, which is $ 0 $ for a separable pair and of order $ 1 $ otherwise. For the multiplicative test, writing $ t_1 = f H_{ij} $ and $ t_2 = (\partial_i f)(\partial_j f) $, the numerator $ t_1 - t_2 $ of $ \partial^2 \ln|f| / \partial x_i \partial x_j $ is a difference of two nearly equal terms exactly when the pair separates, so we score the relative cancellation, the median of $ |t_1 - t_2| \, / \, (|t_1| + |t_2|) $, which is again $ 0 $ when the pair separates. We score the relative cancellation rather than the magnitude of the second derivative itself because that magnitude carries the error of the network's derivatives at the same size as a genuinely coupled pair, leaving no signal to threshold. Points at which $ |f| $ is near zero are excluded from the multiplicative test alone, as $ \ln|f| $ is undefined there, and we attempt no multiplicative decomposition when fewer than $ 8 $ points remain. A grouping of the variables is the set of connected components of the graph of the pairs scoring above a threshold. We enumerate the groupings that a ladder of thresholds produces, discard those from which the network cannot reconstruct its own outputs before any beam search is spent on them, and keep the one whose combined formula attains the highest $ R^2 $ on the held-out split of Section~\ref{subsec:inferenceMethods}.

\subsection{One-shot LLM prompt}
\label{app:oneshot-prompt}

The one-shot LLM method issues one chat completion per task, with the fixed system prompt below and the user prompt after that. Angle brackets mark the per-task substitutions, and the fitting referred to in the prompt is done with BFGS.

\begin{lstlisting}[style=prompt]
You are an exceptional symbolic regression assistant. You analyse numerical
relationships between variables and infer the closed-form mathematical formula
that generated them. You answer with a formula and nothing else.
\end{lstlisting}

\begin{lstlisting}[style=prompt,caption={The one-shot user prompt template.},label={lst:oneshot-prompt},captionpos=b]
Find the mathematical formula for <output description> (<output>) as a function
of the following input variables:
  <variable 1>: <variable 1 description>
  ...
  <variable m>: <variable m description>

Below are 30 observed samples (columns: <variable 1>, ..., <variable m>,
<output>):

  <30 rows of numbers, one per sample, drawn from the training split>

Infer the formula that maps the inputs to the output. It must hold for all the
data above, and it must be physically/mathematically plausible, not a fit of
convenience.

Rules for your answer:
  - Use only these variables: <variable 1>, ..., <variable m>.
  - Use only these operations and functions: +, -, *, /, ** (power), exp, log,
    sqrt, sin, cos, tan, arcsin, arccos, arctan, sinh, cosh, tanh, abs, and the
    constant pi.
  - Write every unknown numeric coefficient as a placeholder symbol c0, c1, c2,
    ... (at most 10). Their values are fitted to the data afterwards by a
    numerical optimizer, so do NOT guess numbers: give the STRUCTURE, e.g.
    c0*x1*exp(-c1*x2) + c2.
    Structural exponents may stay literal (x1**2, sqrt(x1)).
  - Every placeholder must be multiplied/added into the formula somewhere it can
    matter; do not write a placeholder you do not need.
  - Use Python/SymPy infix syntax on a single line, e.g. c0*x1*exp(-c1*x2) + c2.
  - Output ONLY the right-hand side of the expression for <output>. No
    explanation, no units, no code fences, no '=' sign.
\end{lstlisting}

\subsection{Operator sets available to each method}
\label{app:operatorSets}

The methods we evaluate are not given the same set of operators, so we state here what each one may use.

Our transformer emits only the tokens of the grammar $ \cG $ of Section~\ref{subsec:dataGeneration}. No operator outside $ \cG $ can appear at any point of the decoding.

The one-shot method instructs the model to use $ + $, $ - $, $ \times $, $ / $, exponentiation, \texttt{exp}, \texttt{log}, \texttt{sqrt}, \texttt{sin}, \texttt{cos}, \texttt{tan}, \texttt{arcsin}, \texttt{arccos}, \texttt{arctan}, \texttt{sinh}, \texttt{cosh}, \texttt{tanh}, \texttt{abs} and the constant $ \pi $ (Listing~\ref{lst:oneshot-prompt}). That list covers $ \cG $ --- the unary operators $ S+\!+ $, $ S-\!- $, $ 1/S $, $ S^2 $ and $ S^3 $ are expressible through $ + $, $ / $ and exponentiation --- and adds $ \sinh $ and $ \cosh $, which $ \cG $ does not contain. The parser that reads a reply admits more than the prompt offers: it also accepts \texttt{asinh}, \texttt{acosh}, \texttt{atanh}, \texttt{cbrt}, \texttt{log10}, \texttt{atan2}, \texttt{sign} and the constant $ e $. We admit those because the models write the same function in several spellings and we would otherwise discard a reply over its notation. A reply is rejected when it applies a function outside that set, introduces a variable the task does not have, carries more than the $ 10 $ constants we fit, or evaluates to a non-finite or complex value.

The LLM-SR algorithm states no set of operators. It prompts for the body of a Python function with the numerical library NumPy in scope and evolves that program~\citep{shojaee2025llmsr}, so any function of that library and any control flow of the language may appear in a candidate, and a candidate need not be a closed-form formula. A candidate is kept when it returns a finite loss with at most the $ 10 $ constants we fit. We run the algorithm as its authors released it rather than restricting it to $ \cG $, so that our measurements of it remain comparable with the published ones.

Each of the two LLM methods therefore has at least the operators of $ \cG $ available to it, and LLM-SR is not confined to closed-form formulae. Any effect this has on the comparison favors them over our transformer.

\subsection{Best-seed comparison on LSR-Transform}
\label{app:bestseed}

\begin{figure*}[t]
\centering
\includegraphics[width=\textwidth]{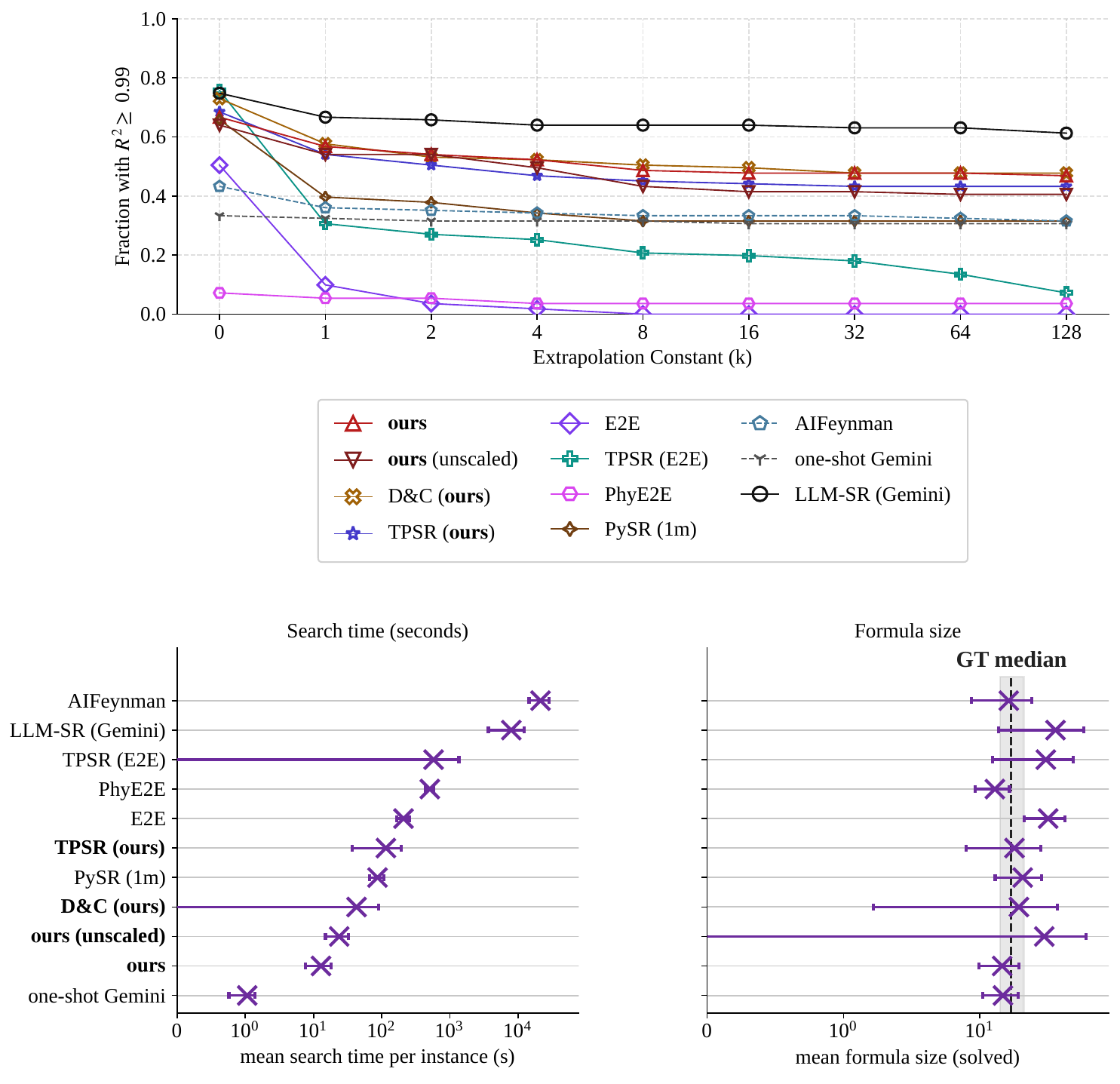}
\caption{Left: solution rate $ \rho_k $ versus extrapolation constant $ k $ on the $ 111 $ LSR-Transform formulae. Right: mean per-formula synthesis time and mean synthesized formula size on the same formulae, with the median size and $ \pm 1 $ standard deviation of the ground-truth formulae shown by the dashed vertical line and the gray region. The LLM-SR with Gemini curve is compared against the single best seed of each other method in the noise-free setting.}
\label{fig:llmsr_extrapolation_time_complexity}
\end{figure*}

Because of API cost we ran LLM-SR with Gemini on LSR-Transform for a single noise-free seed. Figure~\ref{fig:llmsr_extrapolation_time_complexity} compares that run against the seed of each other method with the highest $ \rho_0 $, showing $ \rho_k $ on the left and synthesis time and formula size on the right. Section~\ref{subsec:results} discusses the results.

\subsection{Construction of the extrapolation input ranges}
\label{app:ooddomain}

Shifting the whole input box by $ k \boldsymbol{\delta} $ can leave the domain of the target formula, which would drop the task from the comparison at that $ k $. We therefore choose a per-variable shift. At each $ k $ we first try shifting all variables together; if the target formula does not evaluate to finite values there we advance the variables one at a time, in the order the benchmark lists them, keeping each advance whose outputs stay finite and leaving the remaining variables at their previous shift. A variable that cannot advance in one direction is offered the other, and the direction is settled in a first pass at the largest $ k $ so that a variable does not reverse partway along the $ k $ axis. The per-variable shifts are non-decreasing in $ k $ by construction, so a task never moves back toward its training range as $ k $ grows. This retains all $ 99 $ SRBench and all $ 111 $ LSR-Transform tasks at every $ k $.

\subsection{Formula-trained transformers}
\label{app:fml_transformer_arms}

\begin{figure*}[t]
\centering
\includegraphics[width=\textwidth]{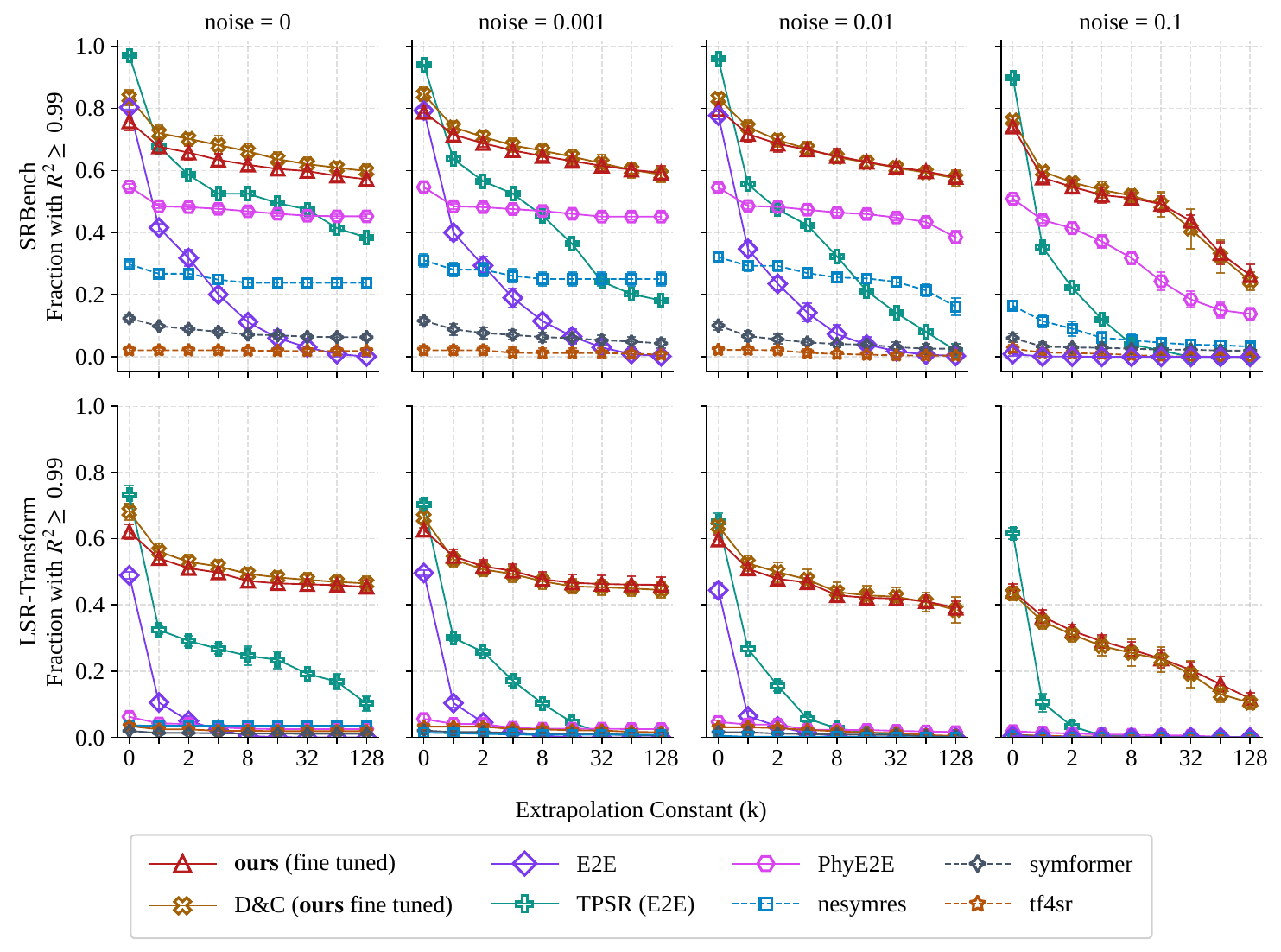}
\caption{Solution rate $ \rho_k $ versus extrapolation constant $ k $ on the $ 99 $ SRBench formulae and the $ 111 $ LSR-Transform formulae, with every formula-trained arm drawn.}
\label{fig:fmltfs_both}
\end{figure*}

\begin{figure*}[t]
\centering
\includegraphics[width=\textwidth]{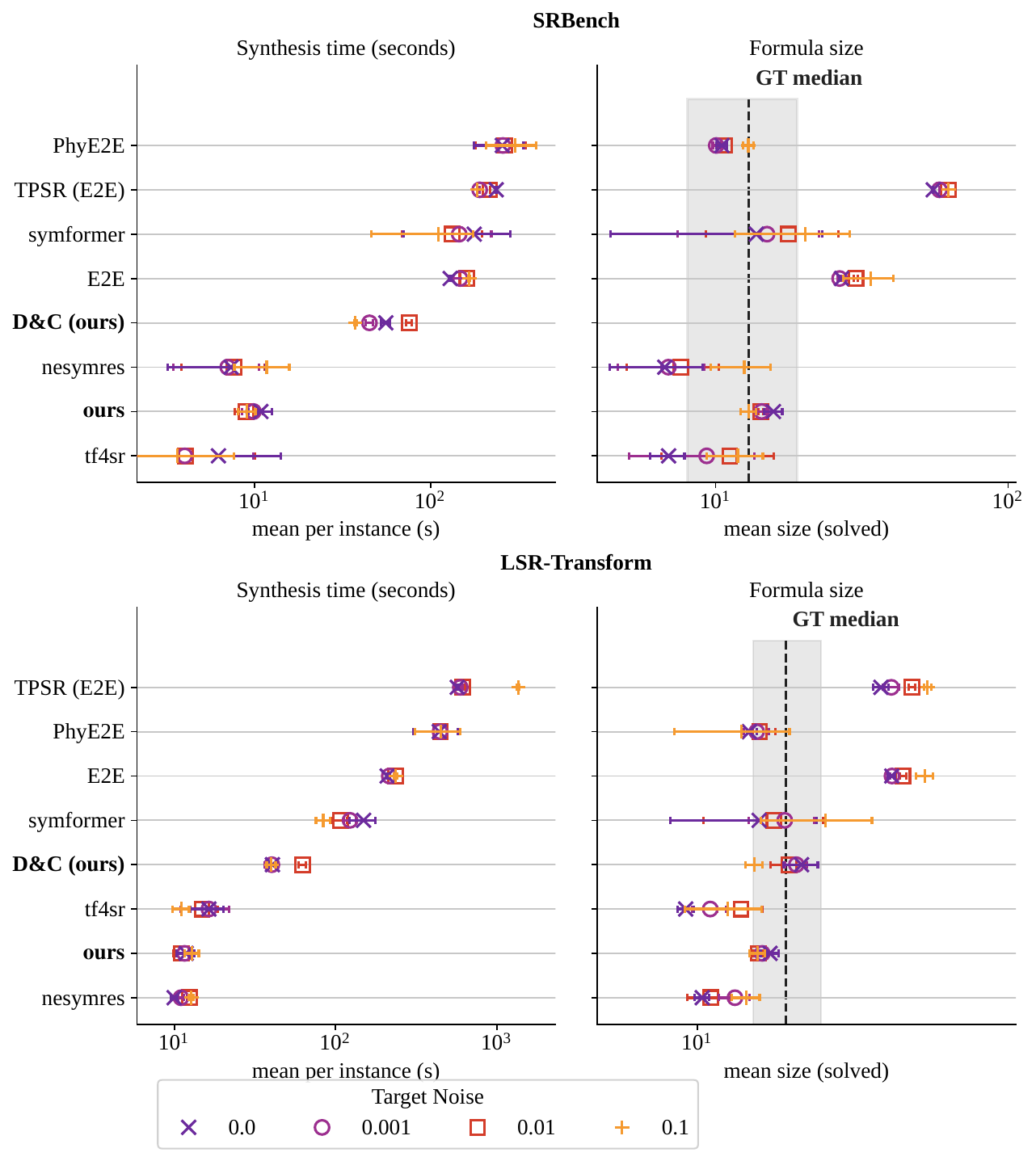}
\caption{Time per formula and predicted formula complexity for SRBench and LSR-Transform.}
\label{fig:fmltfs_both_time_complexity}
\end{figure*}

NeSymReS encodes the data points with a set transformer~\citep{lee2019set} and decodes formula skeletons whose constants are placeholders, which we fit with BFGS for each skeleton in its beam. Its released model accepts at most three input variables. SymFormer decodes each symbol together with the value of its constant and then refines the constants by gradient descent. Its released models accept at most two input variables. TF4SR decodes a formula skeleton whose constants we fit with BFGS. Its released model accepts at most six input variables, all of them positive. Figures~\ref{fig:fmltfs_both} and~\ref{fig:fmltfs_both_time_complexity} show all six methods.

\subsection{Language-trained transformers}
\label{app:llmarms}

\begin{figure*}[t]
\centering
\includegraphics[width=\textwidth]{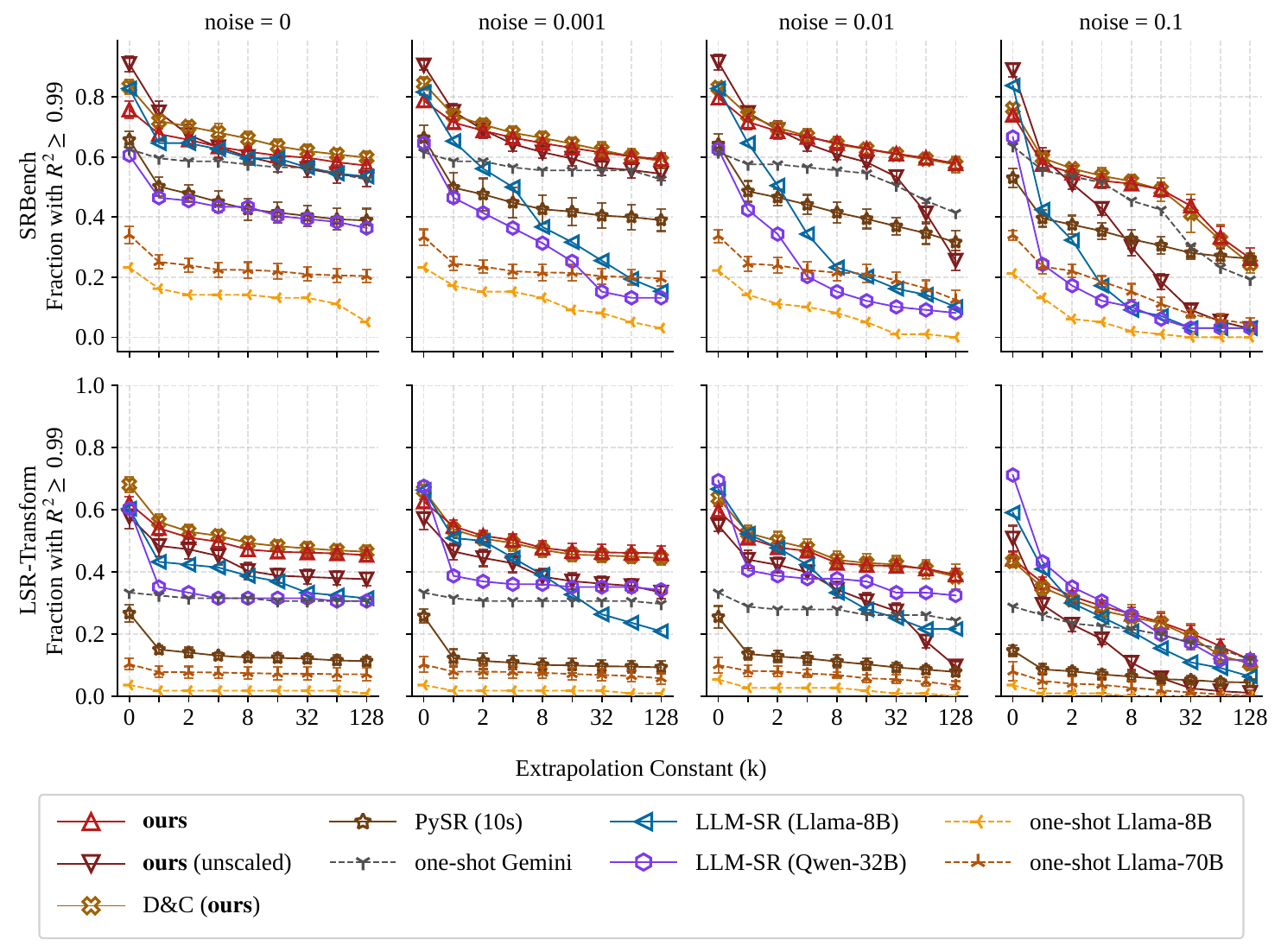}
\caption{Solution rate $ \rho_k $ versus extrapolation constant $ k $ on the $ 99 $ SRBench formulae and the $ 111 $ LSR-Transform formulae, with every LLM arm drawn.}
\label{fig:appx_both_llm}
\end{figure*}

\begin{figure*}[t]
\centering
\includegraphics[width=\textwidth]{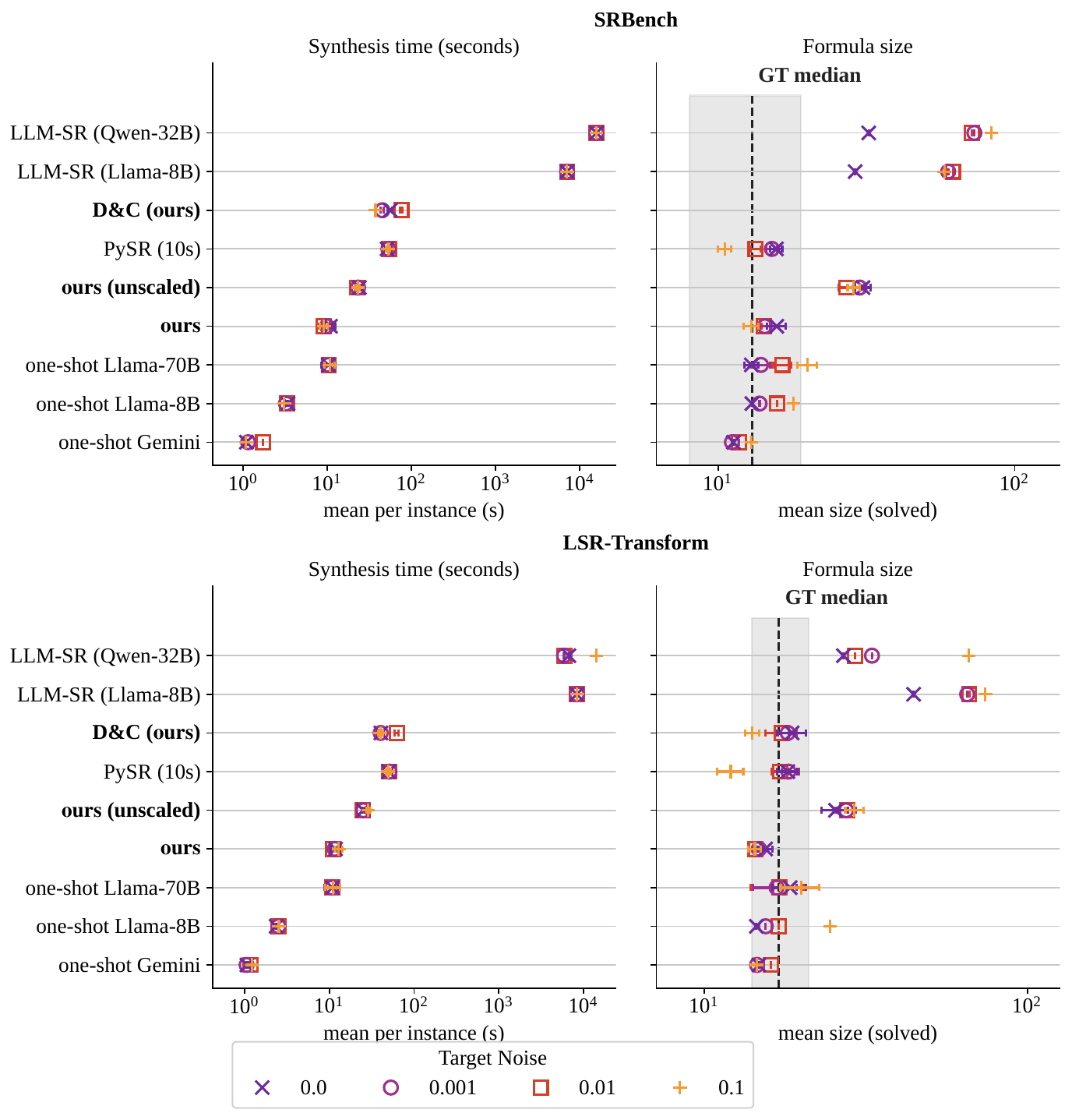}
\caption{Time per formula and predicted formula complexity for SRBench and LSR-Transform.}
\label{fig:appx_both_time_complexity}
\end{figure*}

The body figures draw only one-shot Gemini among the LLMs, to keep the panels legible. Figures~\ref{fig:appx_both_llm} and~\ref{fig:appx_both_time_complexity} redraw the same two benchmarks with every LLM arm we evaluated: one-shot Gemini, one-shot Llama-8B, one-shot Llama-70B, and LLM-SR with Llama-8B and with Qwen. All other arms and all control parameters are as in Section~\ref{subsec:control_parameters}.

\subsection{Control parameter values}
\label{app:controlparameters}

Table~\ref{tab:controlparameters} gives the value of every control parameter that the body of the paper names, in the order in which the body introduces them.

\begin{table}[t]
\centering
\caption{The control parameters from the main text of the paper and their values.}
\begin{tabular}{lll}
\toprule
{\bf Symbol} & {\bf Value} & {\bf Meaning} \\
\midrule
\multicolumn{3}{l}{\em Training data generation (Section~\ref{subsec:dataGeneration})} \\
\quad $ V_\text{max} $      & $ 10 $                    & largest number of variables in a formula \\
\quad $ l $                 & $ 1 $                     & smallest number of nodes in a formula \\
\quad $ r $                 & $ 40 $                    & largest number of nodes in a formula \\
\quad $ \gamma $            & $ 0.99 $                  & decay of the formula size distribution \\
\quad $ N $                 & $ \cU\{50, \ldots, 400\} $ & data points per training instance \\
\cmidrule(r){1-3}
\multicolumn{3}{l}{\em Embedding (Section~\ref{subsec:transformerArchitectureTraining})} \\
\quad $ d_\text{emb} $      & $ 512 $                   & dimension of the data embedding \\
\quad $ d_\text{head} $     & $ 32 $                    & query, key and value dimension per attention head \\
\cmidrule(r){1-3}
\multicolumn{3}{l}{\em Formula synthesis (Section~\ref{subsec:inferenceMethods})} \\
\quad $ N $                 & $ 400 $                   & data points per bag \\
\quad $ B $                 & $ 100 $                   & bags per task \\
\quad $ C $                 & $ 10 $                    & candidate formulae kept per bag \\
\quad $ \theta $            & $ 0.9999 $                & early-stop threshold on beam-search $ R^2 $ \\
\cmidrule(r){1-3}
\multicolumn{3}{l}{\em Empirical evaluation (Section~\ref{subsec:control_parameters})} \\
\quad $ \tau $              & $ 0.99 $                  & threshold for a near-exact match \\
\quad $ k $                 & $ \{0,1,2,4,8,16,32,64,128\} $ & input-shift constants \\
\quad $ n $                 & $ 10{,}000 $              & points per extrapolation data set \\
\quad $ \epsilon $          & $ \{0, 0.001, 0.01, 0.1\} $ & target noise levels \\
\quad $ P $                 & $ 53{,}334 $ / $ 20{,}000 $  & points pooled per task and seed, SRBench / LLM-SRBench \\
\cmidrule(r){1-3}
\multicolumn{3}{l}{\em Further training (Section~\ref{subsec:transformerArchitectureTraining})} \\
\quad $ \epsilon_\text{max} $ & $ 0.1 $              & largest training-time target noise level \\
\quad $ p_\text{clean} $      & $ 0.5 $              & share of formulae kept noise-free \\
\quad $ \eta_\text{ft} $      & $ 5 \times 10^{-5} $ & peak learning rate \\
\quad $ W_\text{ft} $         & $ 500 $              & warmup steps \\
\bottomrule
\end{tabular}
\label{tab:controlparameters}
\end{table}

\end{document}